\documentclass[letterpaper]{article} 
\newcommand{\AllowTechnicalReportHyperref}{}
\usepackage[preprint]{aaai2027}
\usepackage{newtxtt}
\usepackage[hyphens]{url} 
\usepackage{xurl}
\usepackage{graphicx} 
\def\UrlFont{\rm} 
\usepackage{natbib} 
\usepackage{caption} 
\usepackage{amsmath}
\usepackage{amssymb}
\usepackage{newtxmath}
\usepackage{multirow}
\usepackage{booktabs}
\usepackage{tabularx}
\usepackage{array}
\usepackage{xcolor}
\usepackage{float}
\usepackage{placeins}

\usepackage{tikz}
\usetikzlibrary{shapes.geometric, arrows.meta, positioning, calc, patterns}

\usepackage{listings}
\usepackage{nameref}
\usepackage[hidelinks]{hyperref}
\hypersetup{hypertexnames=false}

\newcommand{\breakablett}[1]{%
    \begingroup\def\UrlFont{\ttfamily}\nolinkurl{#1}\endgroup%
}

\definecolor{codebackground}{RGB}{247,248,250}
\definecolor{codeborder}{RGB}{205,210,218}

\title{VoxelSage: Tool-Augmented 3D CT Analysis and Simulator-Shielded Sequential Resection Planning for Liver Tumors}
\author{
Binghong Qian\textsuperscript{\rm 1}, 
Xuanhe Liu\textsuperscript{\rm 1}, 
Yifan Xing\textsuperscript{\rm 1}, 
Wenjie Deng \textsuperscript{\rm 1}, 
Jian Wu \textsuperscript{\rm 1}, 
Haochao Ying \textsuperscript{\rm 1}
}

\affiliations{
\textsuperscript{\rm 1}Zhejiang University\\
}

\begin{document}
\maketitle
\thispagestyle{plain}
\begin{abstract}
Preoperative liver-tumor assessment requires segmentation, physical-space measurement, visual evidence, and resection planning from the same three-dimensional CT volume. Existing tools often handle these steps separately, while language models cannot reliably compute physical measurements from CT. To provide an integrated workflow, we present VoxelSage, a multi-modal system for two- and three-dimensional visualization, liver-tumor analysis, and preoperative resection planning. Its dual-port architecture separates language-model orchestration from image computation: Port A interprets requests and selects skills, while Port B applies them to CT volumes and segmentation masks and returns structured results. Keeping physical measurements in Port B prevents the LLM from computing them directly and reduces the risk of fabricated numerical results. Eight built-in skills support quantitative analysis, visual evidence generation, three-dimensional reconstruction, segmentation refinement, and sequential resection planning; user-defined skills can extend these functions. For sequence planning, a behavior-cloned neural ranker orders candidate resection targets, while a simulator-based shield checks them against predefined constraints. Across 256 unseen simulator scenes, this approach reduced mean simulated time from 34.274 to 33.388 min (0.886 min, 2.59\%) and mean simulated blood loss from 300.847 to 183.852 mL (116.995 mL, 38.89\%) relative to a deterministic baseline. These results demonstrate system integration and simulator-level performance, not clinical efficacy or safety. The public implementation is available at \url{https://github.com/ZJUMAI/VoxelSage}.
\end{abstract}


\section{Introduction}

\begin{figure*}[t]
    \centering
    \includegraphics[width=1\textwidth]{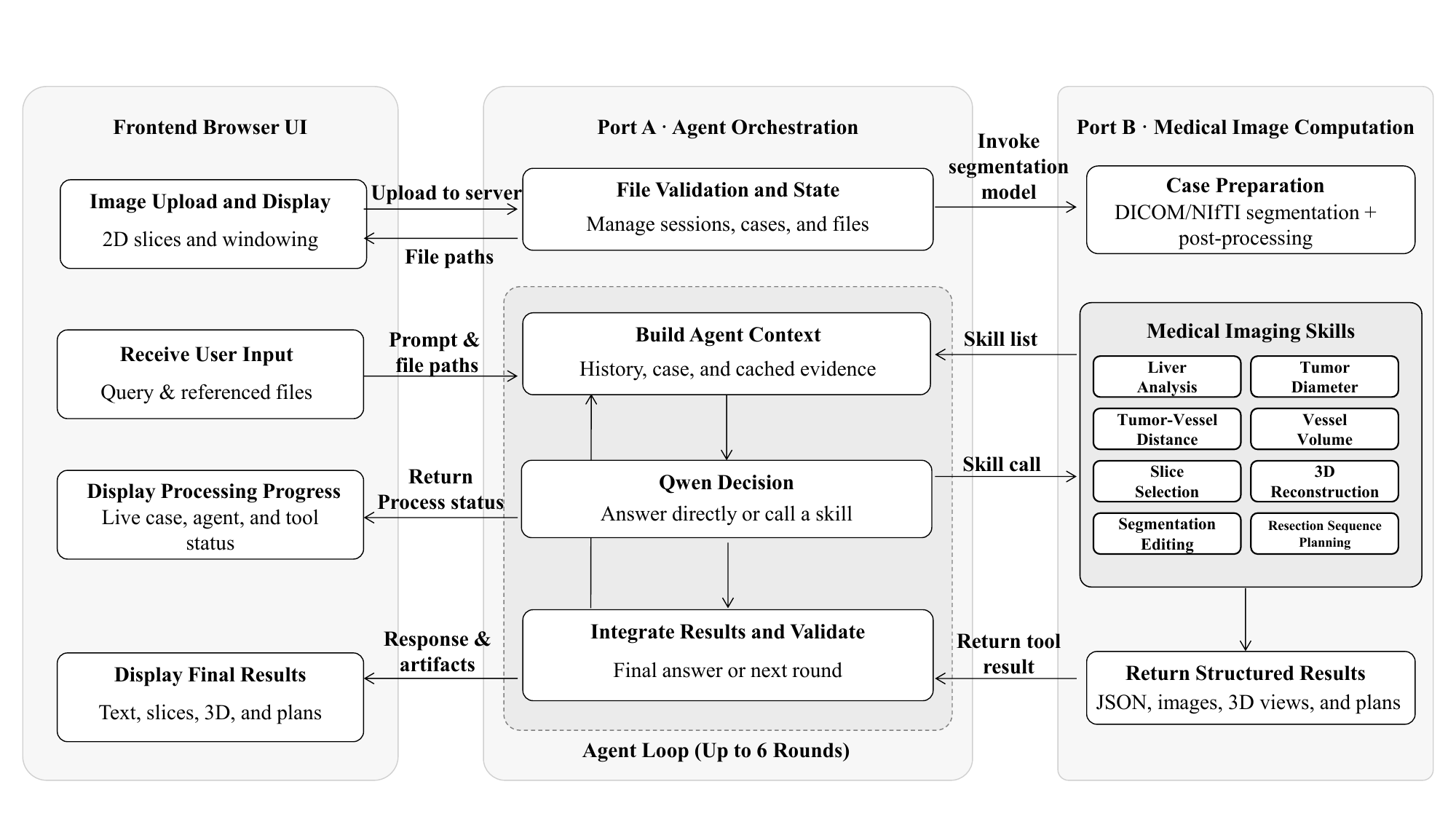}
    \caption{Dual-port collaboration workflow connecting the browser, Port A agent orchestration service, external language model, and Port B medical image computation service.}
    \label{fig:system_architecture}
\end{figure*}

Preoperative liver-tumor assessment requires more than detecting tumors in CT. Information from the full CT volume must be combined to quantify tumor burden and assess vascular relationships, resection margins, and preserved liver tissue. For these findings to support resection planning, the CT volume, segmentation masks, measurements, and visual evidence must remain linked to the same case and physical coordinate system.

Existing methods address individual parts of this task. VISTA3D, TotalSegmentator, and BiomedParse provide automatic or prompt-guided medical-image segmentation \cite{He2025VISTA3D,Wasserthal2023TotalSegmentator,Zhao2025BiomedParse}; CT-CLIP connects volumetric CT with text \cite{Hamamci2026CTCLIP}; and CT-Agent and 3DMedAgent organize specialized tools and selected evidence for multi-step CT reasoning \cite{Mao2026CTAgent,Wang2026ThreeDMedAgent}. IDEAL outlines a liver-surgery decision-support framework that brings together multimodal evidence, three-dimensional anatomy, functional information, and explicit constraints \cite{Wang2026IDEAL}. However, these methods do not provide one case-centered workflow that connects image preparation, image-based measurement, visual evidence, interactive correction, and resection planning. This gap motivates a system that coordinates specialized tools while keeping every result tied to its source data.

To address this need, we present VoxelSage, an online multimodal system for two- and three-dimensional visualization, liver-tumor analysis, and preoperative resection and sequential planning. To separate language-based coordination from computations that require image geometry and to keep the system extensible, VoxelSage uses a dual-port architecture: Port A interprets user requests, maintains the case context, selects skills, and integrates returned evidence, whereas Port~B performs segmentation and applies the selected skills to the CT volume and segmentation masks. Figure~\ref{fig:system_architecture} summarizes this reasoning--execution loop and the data exchanged between the browser, Port A, and Port B.

The main contributions of this work are summarized as follows:

\begin{enumerate}
    \item \textbf{A dual-port architecture that separates task coordination from image computation.}
    To prevent the language model from generating physical-space measurements directly, Port A handles request interpretation and skill selection, while Port B performs the segmentation and the corresponding computation on the CT volume and segmentation masks. To reduce repeated computation and improve efficiency, Port A caches the result received from Port B. For cross-case analysis, Port A also supports explicit file selection, allowing users to specify which cases to include and keeping the analysis focused on the selected files. Eight built-in skills return structured results for analysis, visualization, correction, and planning, or explicit errors on failure.

    \item \textbf{A learned sequential resection planning method with ``shield''.}
  The system uses a bicubic B\'ezier surface to define the planned resection area. On a two-dimensional grid derived from the saved surface, a frozen behavior-cloned neural ranker orders candidate targets, while a simulator-based ``shield'' excludes candidates that would lead to excessive simulated blood loss or violate other predefined constraints.
\end{enumerate}

To evaluate the online system and the learned controller, we use two evaluation settings. An illustrative single-case demonstration shows the end-to-end online workflow, while a three-case functional matrix using public colorectal liver metastasis CT data verifies the online data flow and Port B skill interfaces. A frozen two-dimensional simulator measures the effect of neural target ordering under controlled assumptions. The image-analysis interface uses only liver, tumor, and vessel masks, and it does not use information about whether a tumor is primary liver cancer or a metastasis from another organ. Therefore, evaluation on colorectal liver metastasis cases alone does not establish that the system generalizes to primary liver cancer or to tumors with other origins. The reported evaluation supports engineering integration and simulator-level behavior only, clinical performance and surgical safety require future validation.

\section{Related Work}

\subsection{Three-Dimensional Medical Image Segmentation}

Segmentation is the geometric foundation for CT-based liver assessment: downstream volume, diameter, distance, and surface-reconstruction results are only meaningful when the liver, lesions, and vessels share a reliable spatial reference system.  Recent general-purpose approaches have substantially broadened the structures and interaction modes available to a medical-imaging pipeline.  VISTA3D provides a unified framework for three-dimensional segmentation with automatic and prompt-guided use cases \cite{He2025VISTA3D}; TotalSegmentator provides broad multi-structure CT segmentation \cite{Wasserthal2023TotalSegmentator}; and BiomedParse explores text-conditioned segmentation, detection, and recognition across biomedical images \cite{Zhao2025BiomedParse}.  These models make it practical to obtain masks for analysis without training a distinct segmentation model for every structure or downstream task.

Their outputs, however, are not themselves a complete preoperative-analysis system.  In particular, a multi-label segmentation volume must still be normalized to a common image geometry, converted into lesion instances, checked for consistency across vessel-mask versions and connected to physical-space measurements and visible evidence.  Our work adopts VISTA3D as the default backend for the current segmentation pipeline while retaining interfaces to alternative backends.  Rather than proposing a new segmentation model, we focus on making segmentation a traceable case-level input to deterministic quantification, interactive visualization, and experimental planning.  This distinction is important because the present system demonstrates workflow integration, not a comparative claim of segmentation superiority.

\subsection{Multimodal Medical Agents}

Multimodal medical models increasingly connect images, reports, and question answering, but three-dimensional CT remains difficult to present directly to a general-purpose model.  CT-CLIP demonstrates that volumetric CT features can be aligned with clinical text \cite{Hamamci2026CTCLIP}.  Tool-augmented medical agents address this challenge from a different perspective. CT-Agent supports chest CT question answering and report generation through anatomy-specific tools and global--local token compression, illustrating how tool organization and structured representations can reduce the burden of full-volume reasoning \cite{Mao2026CTAgent}. 3DMedAgent further emphasizes organ-aware memory, selective evidence acquisition, and the reuse of heterogeneous tool outputs for general CT question answering \cite{Wang2026ThreeDMedAgent}.

These efforts motivate the use of a language model in our own system as an orchestrator rather than as a substitute for geometric computation.  Building on this tool-augmented perspective, the present system focuses on an online liver-analysis workflow: a case identifier binds the CT, masks, measurements, images, and three-dimensional assets, while a dedicated imaging service executes the requested operations. Within this workflow, a reported volume, diameter, or tumor--vessel distance is returned together with its associated mask and affine basis, rather than relying on the language model to infer the result from conversational context alone. Our contribution is to organize language-guided skill selection, image-based measurement, evidence review, interactive correction, and resection planning into a case-centered workflow for liver CT analysis.

\subsection{Liver Surgery Planning}

Liver surgery planning requires more than lesion localization. Candidate treatment or resection strategies must be considered in the context of three-dimensional tumor distribution, intrahepatic vessels, safety margins, retained parenchyma, and patient-specific clinical information. A systematic review found that three-dimensional liver reconstruction is used mainly for planning complex or major resections, but also highlighted heterogeneous and predominantly non-randomized evidence \cite{Banchini2024Liver3D}. Future liver remnant volume and function are central clinical quantities that remain outside our current geometric planner \cite{Urraro2025FLR}. IDEAL describes a framework that brings multimodal evidence, three-dimensional anatomy, functional reserve, and explicit safety constraints to support precision liver-surgery decision making \cite{Wang2026IDEAL}. This line of work establishes the value of integrating image-derived anatomy with constraints that surgeons can inspect and discuss, instead of treating a segmentation result as the final planning output.

Deformable B\'ezier surfaces with distance-map margin visualization and NURBS-based virtual resections are established approaches to liver-resection planning \cite{Palomar2017BezierResection,dAlbenzio2024NURBSResection}. Our contribution combines automatic candidate generation and margin refinement with user-confirmed, simulator-shielded target ordering. Starting from the liver, tumor, portal-vein, and hepatic-vein masks, it exposes tumor-distance, vessel-proximity, area, and curvature summaries in a browser scene. It does not model liver function, segmental perfusion, deformable anatomy, instrument mechanics, or postoperative outcomes; nor does it rank treatments as a clinical decision system.

\subsection{Human-in-the-Loop Medical Systems}

Human oversight is especially important when an imaging output can influence interpretation or planning. In this setting, useful collaboration requires more than showing a model prediction: users need access to the evidence behind a measurement, a means to correct or reject an intermediate result, and a clear record of what was accepted. Tool-augmented CT systems such as CT-Agent and 3DMedAgent provide relevant precedents for evidence selection, structured intermediate results, and question-driven tool use \cite{Mao2026CTAgent,Wang2026ThreeDMedAgent}. DECIDE-AI emphasizes transparent reporting of early clinical evaluation, human factors, and workflow effects \cite{Vasey2022DECIDEAI}; those requirements frame future evaluation rather than evidence already supplied by this technical report.

We operationalize this principle at two explicit boundaries.  First, segmentation edits are user initiated and preserved as affine-aware case artifacts, allowing subsequent measurements and visualizations to identify the data on which they depend.  Second, a resection candidate may be inspected and edited through its $4\times4$ control grid, but downstream sequence planning accepts only a surface that the user has explicitly saved.  Edits invalidate older saved states and paths rather than silently propagating them.  These mechanisms do not establish clinical safety or usability; they define an inspectable research workflow in which the system proposes computational evidence and experimental geometry while the human retains modification and confirmation authority.

\section{System Overview}

VoxelSage is a multimodal-agent system for three-dimensional (3D) liver CT analysis. Given a liver CT image and a natural-language request, our agent interprets the request, selects and calls the required analysis skills, and returns a response based on computation evidence. The current implementation and evaluation use public cases from the TCIA Colorectal-Liver-Metastases collection.

\subsection{System Architecture}

To integrate existing analysis tools into a unified workflow and keep the system extensible for new analysis functions, we design a dual-port architecture, which separates general agent functions and specialized medical-image computation in two different ports, one for agent orchestration and the other for computational skills registration. The architecture in Figure~\ref{fig:system_architecture} shows that our work includes a browser workspace and two ports. The browser provides a lightweight and user-friendly agent interface for main dialogue and interactive functions like segmentation correction. Port A handles the task of agent orchestration. It interprets user requests, maintains case context, and coordinates language-model reasoning with skill calls to Port~B. Port~B is the medical-image computation end, with different CT analysis skills registered. When receiving skill calls, it performs the requested deterministic computation and returns structured results as evidence to Port~A for further reasoning.

\subsection{Overall Workflow}

To support case preparation, on-demand skill use, and case-specific evidence flow within the dual-port architecture, the system follows the workflow below: First, in the browser interface, users upload DICOM series or NIfTI volumes that they want to analyze. After receiving the file path, Port~A assigns a unique case id to every case and initializes the corresponding case context, and sends the file paths to Port~B for case pre-processing and pre-analysis. Then, Port~B prepares and segments the CT data and returns the resulting files, status, and basic analysis results to Port~A. After case preparation, the user can submit a natural-language question with optional references to available files.  Based on users' input, case contexts, and available tool list sent by Port~B, Port~A builds the current agent context, and sends it to the language model to reason whether available case evidence is sufficient to answer user's question. If sufficient, the model prepares a response directly; otherwise, it selects appropriate skills and prepares necessary parameters to get more evidence through medical-image computation. Later, the skill calls are further arranged via Port~A and sent to Port~B. In Port~B,  the target case is processed by selected skills, and then structured results together with relevant artifacts, such as images, three-dimensional views, or planning outputs, are returned to Port A. Port A integrates the returned evidence into context, then send back to the language model to determine whether produces the final response or starts another reasoning round (sending skill calls to Port~B, collecting evidence returned from Port~B, and reasoning with evidence to determine whether to produce the final response) when additional evidence is required. To prevent infinite loop, the agent loop is limited to six rounds. If exceeded, the agent will collect logs from both ports to show failure reason. Throughout the workflow, for giving user feedback of execution progress, the browser displays processing status. 

\subsection{Browser Workspace}\label{sec:browser-workspace}

The browser workspace is the interaction and presentation layer for upload, task monitoring, dialogue, evidence review, editing, and confirmation. It displays case-linked structured results and artifacts returned through the architecture in Figure~\ref{fig:system_architecture}; the browser does not compute measurements or alter the source CT data; it sends saved user edits to the backend, which updates the corresponding segmentation or planning artifacts.

After upload and validation, Port A assigns the DICOM series or NIfTI volume a \texttt{case\_id} and submits case preparation. The frontend reports processing or failure state and loads the returned CT, masks, and visual artifacts. The interface displays not only the final answer but also the intermediate evidence that supports it: per-lesion measurement cards, ranked axial slices with contour overlays, and the interactive 3D scene. This arrangement lets the user inspect the case-linked evidence of each quantitative claim before acting on it. Matching cases may reuse valid results; changed inputs invalidate dependent artifacts.

\begin{figure*}[t]
    \centering
    \includegraphics[width=0.9\linewidth]{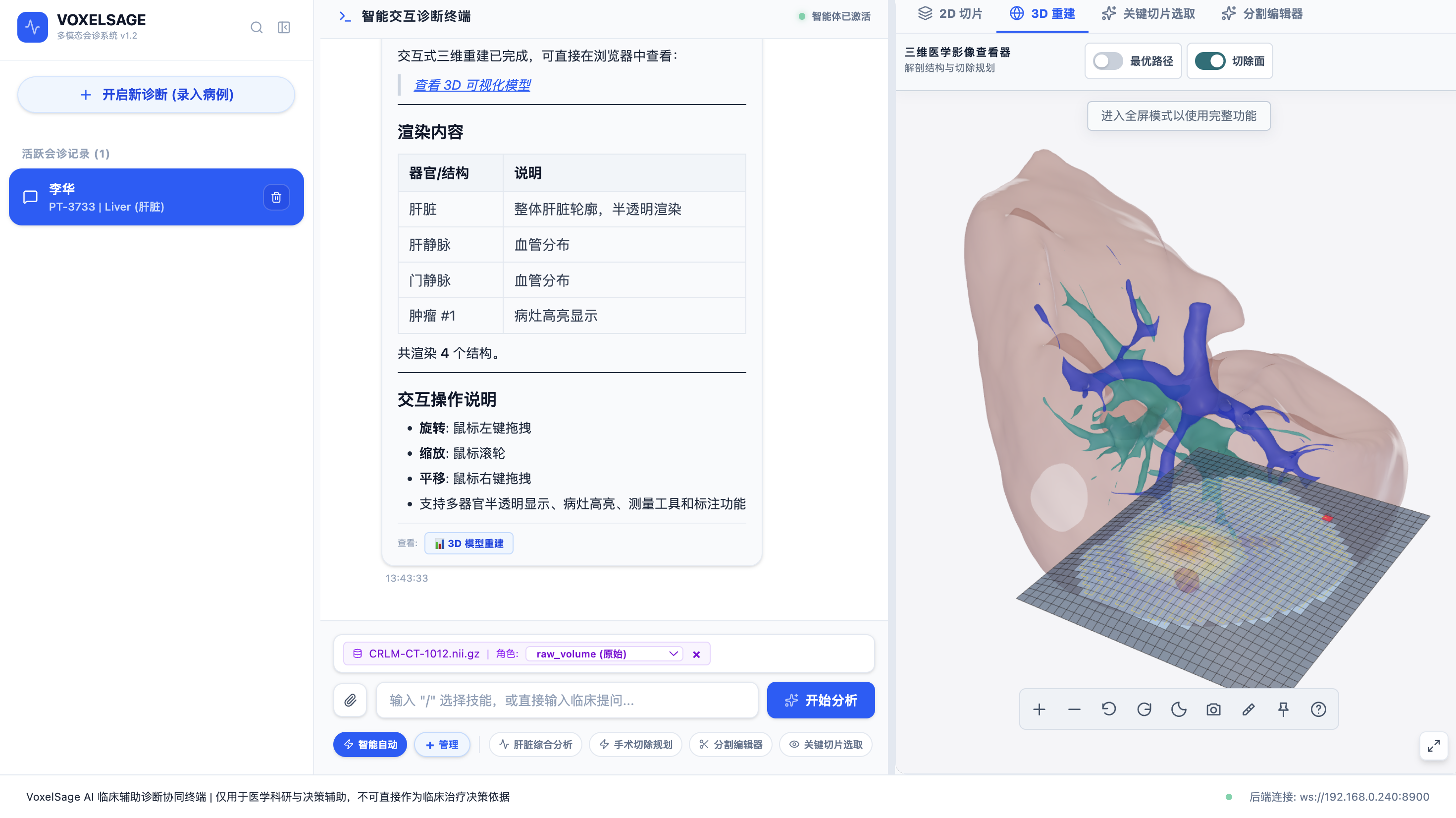}
    \caption{Integrated browser workspace for illustrative case \texttt{CRLM-CT-1012}. The panels combine case management, dialogue, skill execution, and case-linked evidence; the reconstruction panel shows the liver, tumor, portal vein, and hepatic vein in a shared scene.}
    \label{fig:web_overview}
\end{figure*}

Figure~\ref{fig:web_overview} combines case management, dialogue, and skill-specific evidence. Requests carry the active \texttt{case\_id}; Port A discovers skill descriptions through \texttt{/api/skills/list} and submits validated calls to \texttt{/api/skills/run} as described in the orchestration section. The workspace distinguishes the user request, current case state, requested skill, execution status, and returned artifacts. It therefore shows whether a response reused context or triggered computation and presents missing prerequisites or failures explicitly.

Together, case-scoped upload and visible execution state keep conversational requests tied to the data and computation that support each response. This case-level link is the basis for the evidence-review and correction workflows described in Section~\ref{sec:imaging-skills}.The workspace distinguishes three artifact states: generated candidates awaiting review, temporary edits that remain session-local, and explicitly saved artifacts that downstream skills may consume. This distinction lets the interface preserve user control over state-changing operations: segmentation edits take effect only after a deliberate save, and resection planning accepts only a surface that the user has explicitly confirmed. Temporary segmentation edits or unconfirmed resection surfaces cannot enter subsequent computation.

\section{Agent Orchestration}

Port A coordinates the agent loop between user requests and medical-image computation in Port B. The following subsections will illustrate the functions of Port A shown in Figure~\ref{fig:system_architecture} in detail.

\subsection{Case Setup and State Management}

Port A first checks whether an uploaded file is a supported DICOM series or NIfTI (\texttt{.nii.gz}) volume. After the check succeeds, it creates a case context and assigns the case a unique UUID. The case context stores five types of information: the source image, segmentation masks produced during pre-processing, historical medical-image computation results, and file status. The UUID associates source and derived files (e.g. segmented masks) and computational results (e.g. volume of the liver) with the same case, which can effectively avoid misuse. Also, to track progress and handle failures, Port~A records file status, processing status, skill execution states, and errors for each case in the process.

After a skill completes successfully, its structured output (medical-image computation results) is stored in the case context as a case-level cache entry. To reuse a previous computation only for an equivalent request and avoid cache misuse, each entry is indexed by the case UUID, skill name, and input parameters. Port A reuses the output only when all three fields match.

\subsection{Agent Context Construction}

To enable the language model to select an appropriate skill for the current request, Port A first obtains the available skill definition list from Port B. Each definition specifies the skill function, input parameters, prerequisites, and output fields, allowing the language model to understand the function of the skill and what parameters should be provided for skill calling. Port A then combines this skill definition list with the current user request, the historical dialogue, and the case context of the referenced case to construct the current agent context. This context is provided to the language model so that its decision is based on the referenced case and the evidence already obtained.

\subsection{Language Model Decision and Skill Execution}

After receiving the current agent context, Qwen first determines whether the available evidence is sufficient to answer the current user request. If the evidence is sufficient, Qwen generates the response directly. Otherwise, it selects one or more skills from the skill definition list and generates the corresponding skill calls with the required input parameters. Before sending a selected skill call to Port B, Port A checks whether the case context already contains a cache entry with the same case UUID, skill name, and input parameters. If such an entry exists, Port A directly reuses its structured output. Otherwise, it sends a new call to Port B. This check avoids repeated computation for an equivalent request while preventing cache misuse.

When a request requires multiple new computations, Port A sends calls for independent skills to Port B concurrently to reduce waiting time. When one skill requires the output of another skill, Port A sends the calls in sequence so that each skill receives its prerequisite input. Receiving the skill calls from Port A, Port B executes the selected skills on the referenced case and returns structured outputs, generated artifacts, execution states, or explicit errors to Port A. Based on these new information, Port A then rebuilds the context and sends to the language model to judge whether evidence is sufficient. If additional evidence is required, the agent repeats this language model decision and skill execution loop. To prevent infinite agent loop error, the system limits the number of reasoning rounds. If the limit is exceeded, Port A stops further skill calls and asks the language model to review the execution log and return a possible explanation for the unresolved error.

\subsection{Result Integration and Response Preparation}

After Port B returns a successful skill result, Port A adds its structured output and generated artifacts to the case context. The output is also retained as a case-level cache entry; therefore, when a later conversation requests the same skill with the same input parameters for the same case UUID, Port A can reuse the existing result instead of repeating the computation.

Before Port A presents a result to the user, it performs basic range and consistency checks on the returned medical-image computation results. These checks identify possible unit errors or computational anomalies and allow the system to mark the affected result for review. Also, When the workflow fails, Port~B will return the log of error to Port~A. In turn, Port~A adds the error information and a recovery direction to the agent context, allowing the language model to re-enter the reasoning loop and try to fix this error (e.g. adjust the skill calling parameters). However, if the error cannot be recovered, Port A will return a report of execution failure to the user.

\section{Medical Imaging Skills}\label{sec:imaging-skills}

To keep CT analysis extensible without coupling image-processing code to language-model orchestration, we isolate medical-imaging operations in Port B. Port A invokes these operations through schemas exposed from each skill's \texttt{skill.yaml} manifest, while Port B owns case preparation, deterministic computation, and artifact generation. VoxelSage currently provides eight built-in skills and accepts user-supplied packages that follow the registration contract in Appendix~\ref{sec:user-skill-contract}.

For a traceable analysis-and-review loop, the built-in skills are organized into four roles. Measurement skills return mask-derived quantities, evidence-generation skills create inspectable two- and three-dimensional artifacts, interactive correction revises masks under explicit user control, and confirmation-gated sequence planning operates only on a user-saved surface. Table~\ref{tab:builtin_skills} in Appendix~\ref{app:skills} summarizes the eight manifests. The following subsections explain the operating principles of these skills, how the shared preparation stage makes their outputs spatially consistent, how they expose evidence, and how user confirmation controls state-changing operations.

 \subsection{Case Preparation}

\begin{figure}[ht]
    \centering
    \includegraphics[width=.95\linewidth]{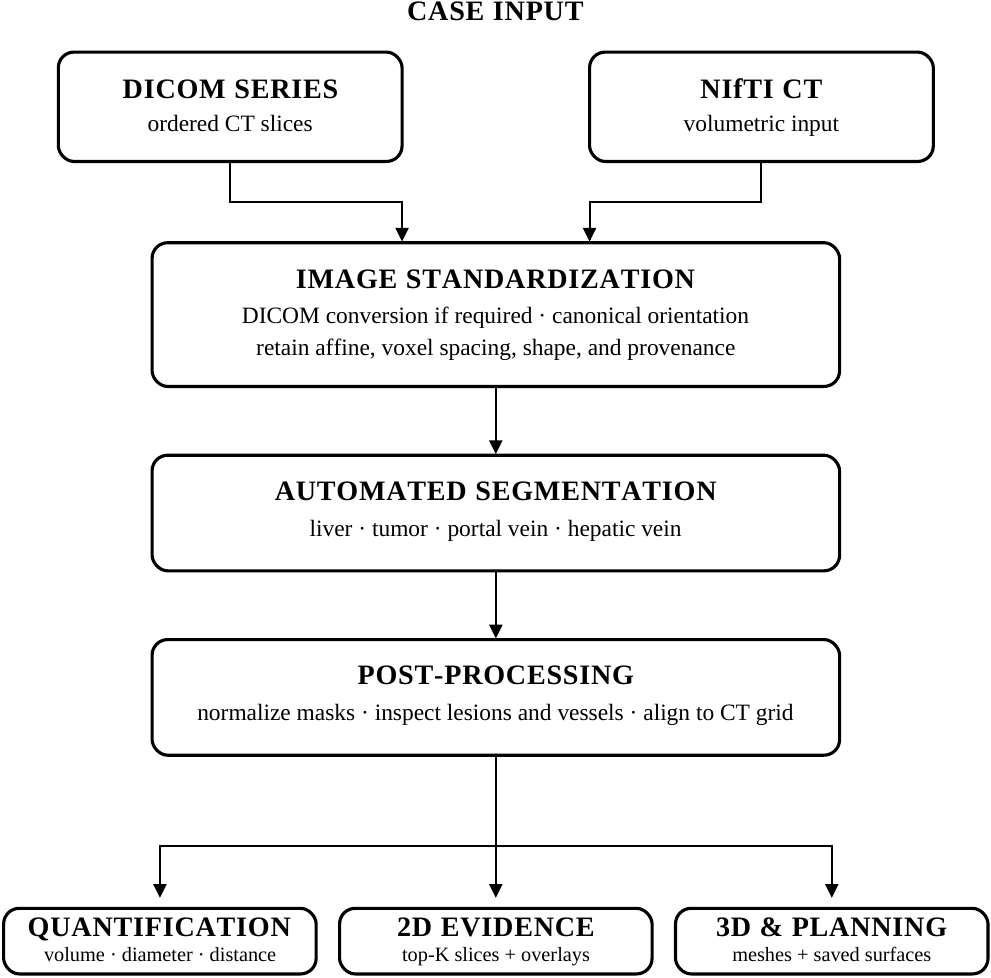}
    \caption{Case preparation and medical-image analysis pipeline. DICOM or NIfTI input is normalized into a canonical case volume. Segmentation and post-processing then produce affine-aligned artifacts that are reused by on-demand quantitative, two-dimensional evidence, and three-dimensional planning skills.}
    \label{fig:medical_image_pipeline}
\end{figure}

To ensure that every downstream skill uses the same patient-space geometry, the preparation service converts DICOM series or NIfTI CT volumes into a common set of case-linked artifacts (Figure~\ref{fig:medical_image_pipeline}). The pipeline has three stages:
\begin{enumerate}
    \item \textbf{Image standardization.}
    To prevent physically misregistered overlays and spacing errors, DICOM series are con verted to case-specific NIfTI volumes, while NIfTI files are used directly. Each volume is loaded in canonical RAS+ orientation and retains its $4\times4$ affine matrix. The mapping $\boldsymbol{x}=\mathbf{A}\boldsymbol{i}$ converts the homogeneous voxel index $\boldsymbol{i}=[i,j,k,1]^{\mathsf T}$ to millimetre coordinates. Masks on another orientation or grid are  canonicalized and, when needed, resampled to the CT grid with nearest-neighbour interpola tion. To serve both display and intensity-sensitive analysis, the service also creates an 8-bit abdominal-window volume at level $40$ HU and width $400$ HU while retaining the unwindowed HU data. The CT path, affine, shape, and output directory are recorded under the case identifier used by later skill calls.
    
    \item \textbf{Automated segmentation.}
    VISTA3D \cite{He2025VISTA3D} serves as the default backend for producing the liver, hepatic-tumor, portal-vein, and hepatic-vein masks required by liver-tumor analysis. These masks are written as NIfTI artifacts. TotalSegmentator \cite{Wasserthal2023TotalSegmentator} is available as an alternative backend.

    \item \textbf{Post-processing.}
    Post-processing gives every skill a canonical mask contract by normalizing backend-specific names and decomposing the aggregate hepatic-tumor mask into three-dimensional connected components. Components below the configurable voxel threshold are not materialized as lesion-specific masks, which keeps likely fragments out of per-lesion geometry. Because segmentation can leave short, artificial breaks in structures that should be connected, vessel optimization constructs a separate, continuity-corrected mask variant for downstream vessel-dependent analysis. Portal- and hepatic-vein masks are skeletonized, and candidate terminal gaps of at most $4$ mm are evaluated. A connection is accepted only if endpoint directions are compatible, its path stays predominantly inside the liver, avoids tumor and the other vessel class, and satisfies radius and added-volume limits. Accepted connections are rasterized as local tubes into separate optimized portal- and hepatic-vein NIfTI masks. These optimized masks are retained as complete alternative artifacts rather than external patch files, while the original masks remain unchanged. An audit report records accepted and rejected links.

    For internally consistent vessel-dependent results, VoxelSage reconstructs and smooths the vessel structure for better visualization effects and analysis convenience when possible.
\end{enumerate}

Together, standardization, segmentation, and post-processing produce the affine-aligned artifacts shown in Figure~\ref{fig:medical_image_pipeline}. Interactive correction remains a separate user-invoked operation, as described below.

\subsection{Quantitative Measurement Skills}

To answer quantitative questions without asking the language model to infer geometry from screenshots, four skills compute structured measurements from the prepared masks and affine. \texttt{liver\_analysis} returns liver and vessel volumes, per-lesion diameters, tumor--vessel distances, and a formatted report in one call. \texttt{tumor\_diameter}, \texttt{tumor\_vessel\_distance}, and \texttt{vessel\_volume} expose the corresponding quantities separately when a focused query does not require the full report. To avoid repeated computation, Port A can reuse relevant fields from the integrated results for focused follow-up questions. Before measurement, the preparation stage can split the aggregate hepatic-tumor mask into separate per-lesion component masks so that each lesion is measured independently. For a binary structure mask $M$, physical volume is
\begin{equation}
V(M)=\left(\sum_{\boldsymbol{i}}\mathbb{I}[M(\boldsymbol{i})>0]\right)
\left|\det\left(\mathbf{A}_{1:3,1:3}\right)\right|,
\end{equation}
where the determinant gives voxel volume in $\mathrm{mm}^3$. Lesion count is therefore based on valid three-dimensional components rather than a single axial slice.

To report diameters in millimetres without assuming isotropic voxels, lesion coordinates are transformed through the affine before the maximum three-dimensional extent is estimated. Sufficiently large components use the maximum pairwise distance among convex-hull vertices. Smaller components use the maximum pairwise distance among all affine-transformed foreground voxel centres, while components below the minimum size threshold are labelled too small for a stable estimate.

To expose potential tumor--vessel proximity, the implementation evaluates a Euclidean distance transform of the vessel complement over the tumor region \cite{Maurer2003DistanceTransform}. The transform uses the three voxel spacings as per-axis sampling distances, giving the minimum foreground voxel-centre distance in millimetres on the aligned orthogonal grid, including anisotropic grids and their rigid rotations. Direct overlap is reported as zero; an empty tumor or vessel mask returns no distance. The result records the vessel-mask variant, minimum distance, and contact flag. Taken together, these measurements provide reproducible evidence derived from segmentation masks rather than independent diagnostic findings.

\subsection{Informative Slice Selection Skill and Two-Dimensional Evidence Review}

The \texttt{slice\_selection} skill provides compact visual evidence without sending an entire CT volume to the conversational model by ranking axial slices and returning three by default. Masks are first aligned to the canonical CT grid. The score combines organ diversity, organ coverage, a lesion bonus, slice centrality, and image entropy, with an additional bonus when lesion and vessel masks co-occur. Candidates are selected in descending score order with a minimum axial separation, relaxed only when too few positive-scoring slices remain to fill the Top-$K$ set. Each result includes the score, a windowed CT PNG, and a radiologically oriented contour overlay.

\begin{figure}[ht]
    \centering
    \includegraphics[width=0.7\linewidth]{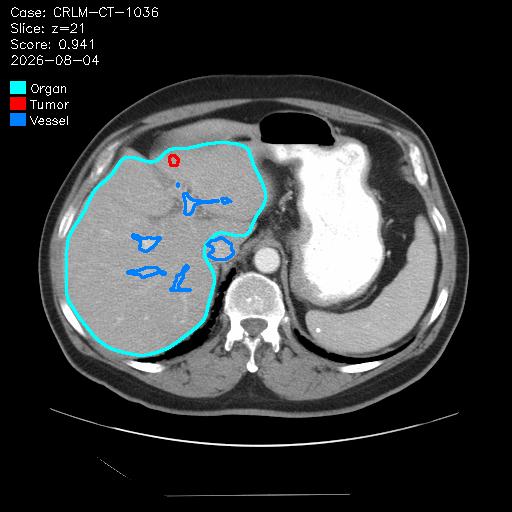}
    \caption{Two-dimensional evidence returned by the \texttt{slice\_selection} skill for illustrative case \texttt{CRLM-CT-1036}. The ranked axial slice shows contours for the liver (cyan), tumor (red), and vessels (blue).}
    \label{fig:slice}
\end{figure}

Figure~\ref{fig:slice} shows one returned overlay. Its colour-coded contours, case identifier, axial index, score, and legend make the selected evidence and its ranking inspectable alongside the companion raw CT image.Structured measurements derived from the same masks—volumes, diameters, and distances—are displayed beside the image rather than embedded in the overlay, so the user can review the numerical evidence and its visual basis side by side without obstructing the underlying anatomy.

\subsection{Interactive Segmentation Correction Skill}\label{sec:seg-correction}

\begin{figure}[ht]
    \centering
    \includegraphics[width=0.8\linewidth]{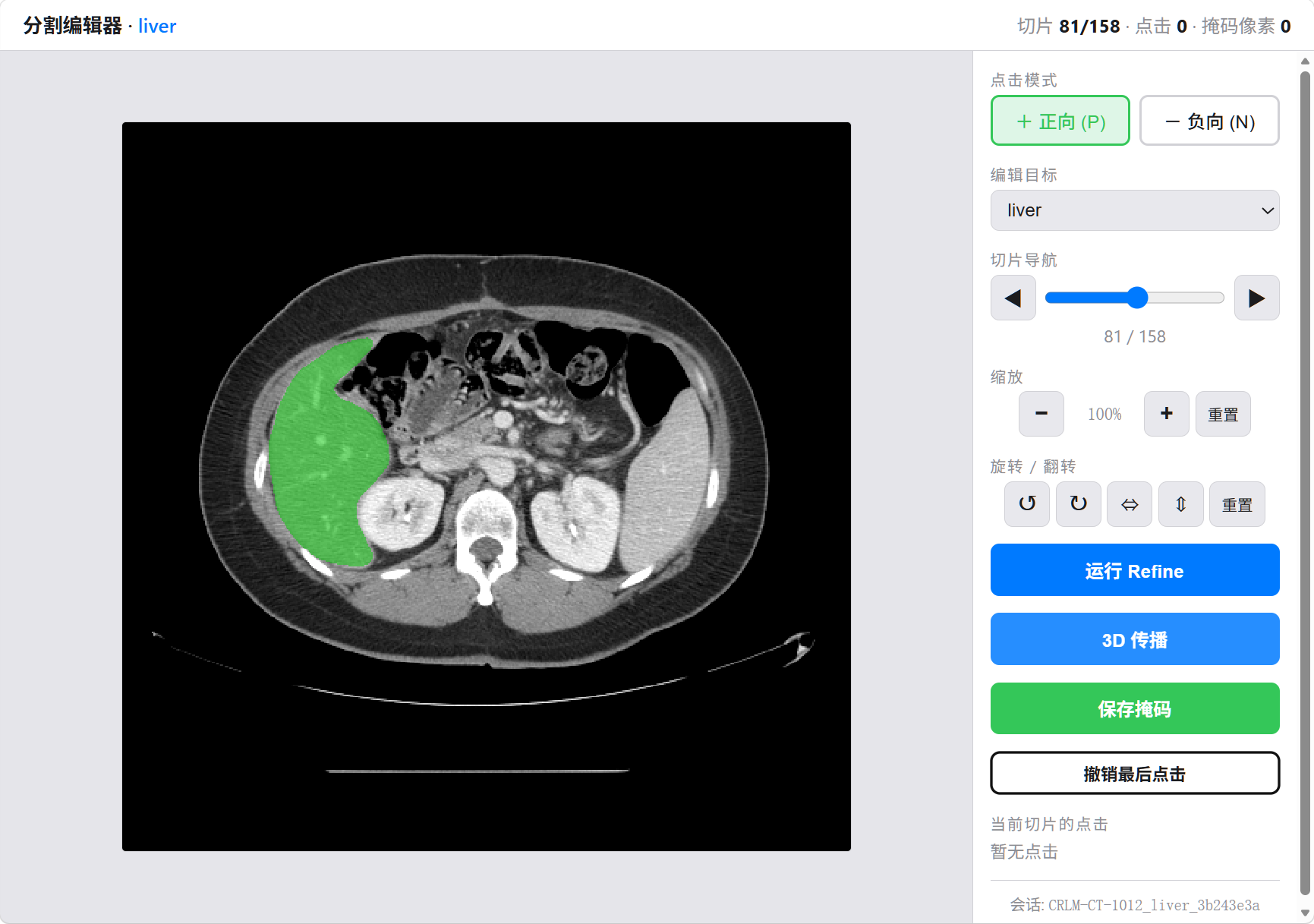}
    \caption{Case-specific editor for interactive segmentation correction, showing the selected structure mask overlaid on an axial slice and the controls (slice navigation, structure selection, positive/negative point prompts, view transformation, single-slice refinement, bounded 3D propagation, undo, and save). The displayed mask remains a working result until the user saves it.}
    \label{fig:segmentation_modification}
\end{figure}

To keep segmentation changes reviewable and explicitly authorized, \texttt{segmentation\_modification} edits a case mask in a working session before any file is replaced (Figure~\ref{fig:segmentation_modification}). The editor supports slice and structure selection, positive and negative point prompts, single-slice refinement, undo, and explicit saving. Optional MedSAM2 propagation can extend accepted edits to a bounded range of adjacent slices \cite{Ma2025MedSAM2}. Until the user saves, other skills continue to read the original mask. Saving preserves the affine, writes the edited mask under the selected structure name, and creates a \texttt{.bak} copy of the previous file so that later skill calls can use the confirmed correction.

\subsection{Three-Dimensional Reconstruction and Interactive Visualization Skill}

To place anatomy and resection proposals in one reviewable frame, \texttt{three\_d\_reconstruction} first builds a self-contained scene from the case masks and then overlays editable B\'ezier surfaces when planning is requested.

\subsubsection{Three-Dimensional Reconstruction}

\begin{figure}[!ht]
    \centering
    \includegraphics[width=0.75\linewidth]{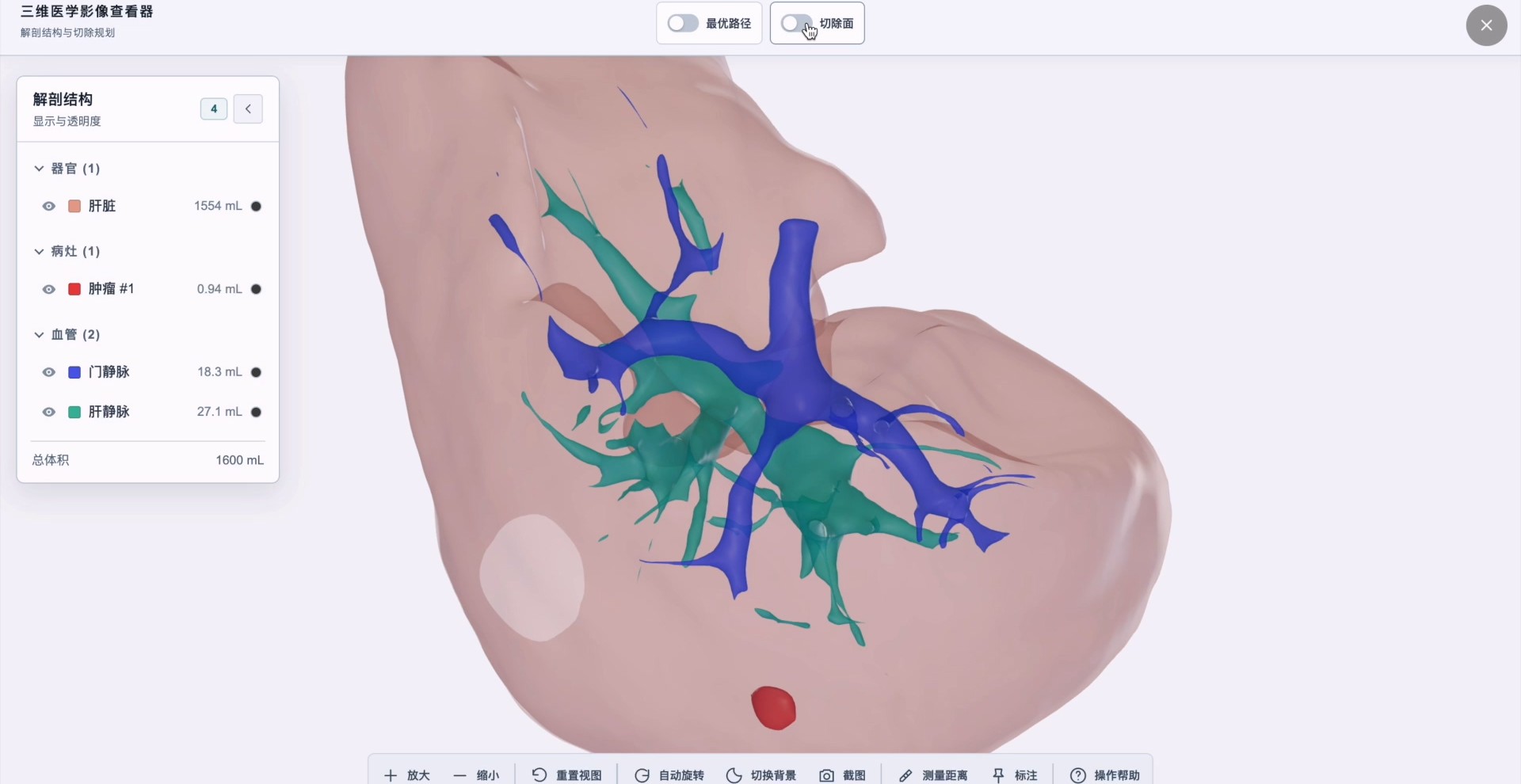}
    \caption{Interactive 3D reconstruction built from the case masks, showing the semi-transparent liver, tumor, portal vein, and hepatic vein in a shared coordinate frame, with structure toggles, navigation, distance measurement, annotation, and mask-derived volumes.}
    \label{fig:three_d}
\end{figure}

The reconstruction supports interactive evidence review by placing liver, tumors, and vessels in a shared Three.js coordinate frame (Figure~\ref{fig:three_d}). Users can adjust visibility, navigate the anatomy, inspect mask-derived volumes, and create interactive distance measurements or annotations in the scene. The viewer also supports screenshot capture and free-form annotations directly on the 3D scene, allowing users to record observations or flag suspicious regions for later discussion. Marching Cubes \cite{Lorensen1987MarchingCubes} extracts each available mask surface, and the case affine maps its vertices from voxel to world coordinates. Optional Laplacian smoothing reduces staircase artefacts. The liver, tumor, portal-vein, and hepatic-vein meshes retain structure labels, colours, mask-derived volumes, and share a common scene-centering offset. Applying that offset to anatomy and later resection surfaces preserves their relative positions, while configurable sampling and downsampling balance mesh detail against browser cost.

\subsubsection{B\'ezier Resection Surface Editing}

When resection planning is requested, the candidate resection surfaces from Section~\ref{sec:resection-planning} are added to the reconstructed scene so that their spatial relationship to the tumor and vascular anatomy remains explicit. The interface provides per-candidate visibility controls, maps displayed-vertex colour to sampled nearest-tumor distance, and exposes the $4\times4$ control grid for editing.

\section{Resection Planning}\label{sec:resection-planning}

To support preoperative review, VoxelSage uses the three-dimensional positions of the tumor, liver, and nearby vessels to generate three diverse candidate resection surfaces. The user can select a candidate and reshape it through a $4\times4$ control grid before saving it. The system then plans an order for moving an ultrasonic-dissector proxy across the confirmed surface. To reduce unnecessary travel and the simulated blood-loss proxy, we train a learned target-ordering model with a simulator-rollout shield. On the frozen 256-scene simulator set, the learned controller reduced mean simulated completion time from $34.274$ to $33.388$ min ($2.6\%$) and mean simulated blood from $300.847$ to $183.852$ mL ($38.9\%$) relative to the serpentine baseline.

\subsection{Resection Surface Representation}\label{sec:resection-surface}

\begin{figure}[!ht]
    \centering
    \includegraphics[width=0.75\linewidth]{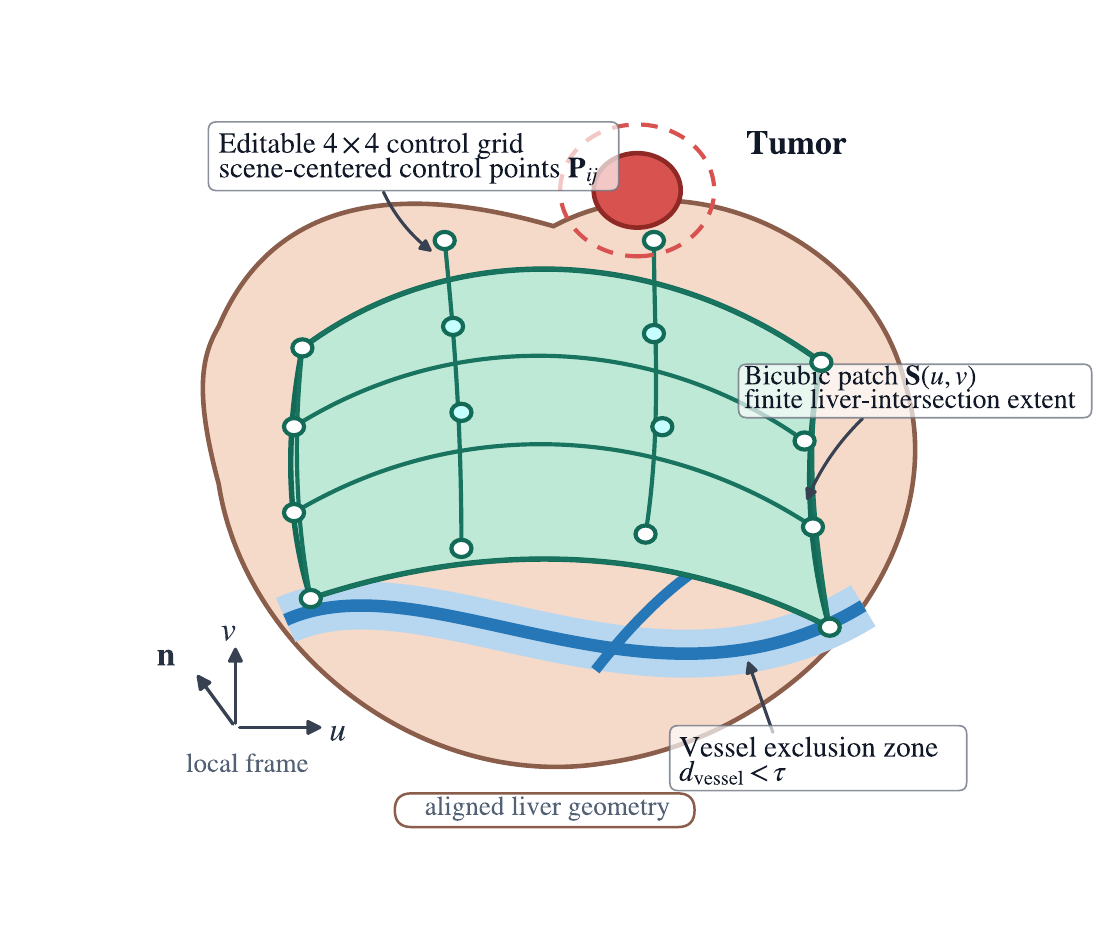}
    \caption{Geometric representation of an editable bicubic B\'ezier resection surface. The $4\times4$ control grid is stored in scene-centered coordinates and mapped back to CT world coordinates through the case center offset. Tumor-margin refinement and the downstream vessel-proximity threshold are shown as distinct geometric checks.}
    \label{fig:bezier_resection_surface}
\end{figure}

To make a candidate smooth enough for editing while using a limited number (16) of control points to define the surface, each resection surface is a bicubic B\'ezier patch \cite{PieglTiller1997NURBS,Farin2002CAGD} controlled by a $4\times4$ grid (Figure~\ref{fig:bezier_resection_surface}). With control points $\boldsymbol{P}_{ij}$, $i,j\in\{0,1,2,3\}$, the surface is
\begin{equation}
\boldsymbol{S}(u,v)=\sum_{i=0}^{3}\sum_{j=0}^{3}B_i^3(u)B_j^3(v)\boldsymbol{P}_{ij},
\qquad (u,v)\in[0,1]^2,
\end{equation}
where $B_i^3$ are cubic Bernstein basis functions.

\subsection{Safety-Margin Constraints}\label{sec:safety-margin}

To enforce a reproducible tumor-side margin during B\'ezier refinement, the planner evaluates signed local clearance at sampled tumor-boundary points. For boundary point $\boldsymbol{x}$ projected to the candidate surface at $(u_x,v_x)$, the tangent-corrected clearance is
\begin{equation}
c(\boldsymbol{x})=
\frac{h(u_x,v_x)-n_x}
{\sqrt{1+\|\nabla h(u_x,v_x)\|_2^2}},
\end{equation}
where $h$ is the B\'ezier height and $n_x$ is the point's coordinate along the reference normal. The gradient is taken with respect to physical in-plane coordinates: $\nabla h=(L_u^{-1}\partial h/\partial u,L_v^{-1}\partial h/\partial v)$, where $L_u,L_v$ are the patch's physical side lengths in millimetres and $u,v\in[0,1]$. If a point projects outside the finite extent of the patch, the candidate does not cover that point, so its clearance is assigned $-\infty$ as an invalid placeholder rather than as a physical distance. The sampled minimum clearance is
\begin{equation}
d_{\min}=\min_{\boldsymbol{x}\in\partial\mathcal{T}}c(\boldsymbol{x}).
\end{equation}

To reject candidates with a local shortfall, the implementation uses a configurable target of $5\,\mathrm{mm}$ and accepts a refined candidate only when $d_{\min}\geq 4.95\,\mathrm{mm}$ (a $0.05\,\mathrm{mm}$ engineering tolerance below the target). This is an engineering parameter rather than a clinical recommendation. The planner also reports the 5th percentile of $c(\boldsymbol{x})$ so that a reviewer can distinguish an isolated minimum from the broader boundary distribution. The colour coding of the resection surface in the rightmost panel of Figure~\ref{fig:web_overview} is a visualization measure based on each displayed surface vertex's nearest-tumor Euclidean distance.

\subsection{Candidate Surface Generation and Evaluation}\label{sec:candidate-generation}

The planner generates candidate B\'ezier surfaces from tumor--liver geometry and several orientations informed by liver and vessel shape. It screens them for tumor coverage, resection extent, detachable-liver support, and vessel exposure, then refines a diverse shortlist toward the configured tumor-margin constraint. The highest-ranked admissible refined surface becomes the primary recommendation.

Two additional surfaces are selected from the unrefined bank to provide different target resection ratios for review. They bypass the primary margin-refinement and case-mode gates and are marked for separate inspection. If no refined surface passes the case-specific gate, the system shows a coverage-qualified fallback with an explicit review flag. These selection rules are engineering heuristics, not clinically validated surgical criteria.

\subsection{Sequential Surface-Coverage Planning with Learned Ordering}
\label{sec:sequential-planning}

After the user saves an edited surface, the planner maps its $4\times4$ control grid back to CT coordinates and divides the liver--surface intersection into approximately $4\,\mathrm{mm}$ cells. The planning task is to order these cells for step-wise coverage by an ultrasonic-dissector proxy. A serpentine scan provides the baseline. Our learned mode keeps the same grid, movement rules, and start condition but learns which eligible frontier target should be visited next; deterministic shortest-path transfer handles movement between targets.

For training and evaluation, the confirmed surface is represented in a reproducible two-dimensional simulator. Vessel cross-sections are retained as separate components: each remains hidden until the surrounding tissue ring has been cut, is then exposed, and is sealed when the path enters it; a boundary component follows the same transition but may be released early if it blocks every remaining non-vessel frontier cell. This preserves a complete coverage path without treating a boundary vessel as ordinary tissue.

The learned model ranks a small set of candidate targets, and a simulator-rollout shield checks them in that order before execution. The first candidate that can complete coverage without exceeding the frozen simulated-blood budget is accepted. Thus, the model changes target order only; the simulator controls travel, clamp timing, exposure, and sealing. Figure~\ref{fig:learned-planner} summarizes training and shielded execution. The simulator omits vessel branching, deformation, perfusion territories, instrument reachability, cutting dynamics, intra-operative registration, and physiological response, so the resulting sequence is a controlled planning proxy rather than a surgical trajectory.

\subsubsection{Planar Simulator and Blood-Loss Proxy}\label{sec:sim-proxy}

The simulator uses a fixed $30\times40$ canvas of $4\,\mathrm{mm}\times4\,\mathrm{mm}$ cells. This cell size is an instrument-scale discretization motivated by the reported $5$-mm UltraCision class, not a measured blade width \cite{Schmidbauer2002Ultracision}. With the reported CUSA-group transection speed of $2.3\,\mathrm{cm}^2/\mathrm{min}$, one cutting or transfer step costs $4.17$ s \cite{Lesurtel2005Transection}. Charging movement through previously cut cells makes inefficient travel visible in the total simulated time.

The simulator repeats a $15$-min clamped and $5$-min unclamped cycle, following a reported range for intermittent clamping \cite{Chouillard2010Clamping}. Simulated blood is zero while clamped, before a vessel component is exposed, and after it is sealed. During an unclamped interval with exposed, unsealed vessels, the rate is
\begin{equation}
Q_t \;=\; \min\bigl(Q_{\mathrm{ref}}(W),\;p_{\mathrm{bleed}}\,\beta(W)\,A_t^{\mathrm{exp}}\bigr),
\end{equation}
where $A_t^{\mathrm{exp}}$ is the exposed area, $p_{\mathrm{bleed}}$ is a fixed scene-level coefficient, and the cap $Q_{\mathrm{ref}}(W)=17W$ mL/min uses the summed hepatic-artery and portal-vein flow scale reported by \citet{Carlisle1992HepaticFlow}. Components spanning at least two cells require three ordinary step times to seal; smaller components require one. The grid size, nominal $70$-kg weight, exposure scale, and sealing times are frozen engineering choices. Consequently, this blood quantity supports within-simulator comparison only and is not a patient-specific estimate.

\subsubsection{Hierarchical Learned Ordering with Simulator-Rollout Shield}\label{sec:sim-controller}

Table~\ref{tab:controllers} defines the controller shorthand before it is used in the formulation and experiments. All six controllers share the same simulator, candidate generator, deterministic transfer, tie-breaking, and clamp schedule, so their differences are confined to target ordering and exact-rollout use.

\begin{table*}[t]
\centering
\small
\caption{Controller definitions. All controllers use the same frozen simulator, legal-target generator, deterministic transfer, tie-breaking, and automatic clamp schedule.}
\label{tab:controllers}
\begin{tabularx}{\textwidth}{@{}lXll@{}}
\toprule
ID & Target ordering & Exact-rollout use & Experimental role \\
\midrule
C0 & Direct deterministic serpentine order & None & Primary reference \\
C1 & Serpentine candidate receives highest priority & Exact admissibility control & Framework control; required to match C0 \\
C2 & Myopic action time, action blood, transfer length, then fixed ties & Exact admissibility control & Non-learned comparator \\
C3 & Corrected depth-one teacher minimizing full-episode time, then blood & All candidates, intrinsic to its objective & Computational teacher reference \\
C4 & Frozen behaviour-cloned macro-target ranker & Lazy exact, in rank order & Primary learned controller \\
C5 & Same frozen ranker as C4 & None & Diagnostic shield ablation \\
\bottomrule
\end{tabularx}
\end{table*}
C2 provides a non-learned ordering under the same candidate set and exact admissibility check as C4. The C4--C2 comparison therefore tests whether learned ranking improves simulated time beyond this specified myopic heuristic; C1 instead checks that adding the shield leaves the serpentine reference C0 unchanged.

\paragraph{Problem formulation.}
To learn only target ordering, we formalize the frozen simulator as a Markov decision process whose low-level transition remains deterministic. The state $s_t$ contains cut cells, current position, hidden, exposed, and sealed component sets, clamping phase and elapsed time, total expected blood loss, and the budget context of spent blood, scene baseline $B_{S,\mathrm{condition}}$, margin $M_B$, and remaining budget. The macro action $a_t$ is a frontier target $g\in\mathcal{F}(s_t)$. On a clone, the transition executes the deterministic shortest transfer followed by that target, and the per-step cost is added time and expected blood $(T_t,B_t)$. Coverage completes when $\mathrm{cut}=D$; illegal actions, stagnation, and time or step limits can terminate an episode unsuccessfully. The shield separately enforces the episode blood budget, whereas unshielded C5 can complete coverage and still exceed that budget. Training and nominal evaluation use the same frozen kernel and automatic $15/5$ clamp schedule, so only frontier ordering is learned.

\paragraph{Candidate generator.}
To expose a small but diverse action set, a deterministic frontier generator returns at most $K=6$ targets $\mathcal{G}_t=\{g_1,\ldots,g_K\}$. It inserts the unique serpentine target first, then cycles round-robin through the other three sources until $K$ unique targets have been collected or the frontier is exhausted:
\begin{enumerate}
    \item the next serpentine target $g^{\mathrm{S}}_t$, labelled \texttt{s\_target}.
    \item one immediately sealable entry for each exposed component, considering larger components first and otherwise preferring entries that require less repositioning and occur earlier in the serpentine order, labelled \texttt{exposed}.
    \item frontier cells next to hidden components, prioritising cells beside a larger total hidden cross-sectional area and otherwise preferring less repositioning and earlier serpentine positions, labelled \texttt{near\_hidden}.
    \item all legal frontier cells, considered from least to most repositioning and with serpentine order breaking ties, labelled \texttt{nearest}.
\end{enumerate}
The capped candidate set does not cover the entire frontier, but it always includes the serpentine fallback and, when available, interleaves exposed, near-hidden, and distance-based alternatives.
Each target feature vector records how the candidate was generated, together with its transfer distance, predicted macro-action costs $(T_t,B_t)$, component area, seal and large-vessel status, and budget context. Because every controller and the shield use this same generator, their comparison isolates ordering rather than candidate-set differences.

\begin{figure*}[t]
    \centering
    \begin{tikzpicture}[
    >={Stealth[length=2.5mm]},
    font=\rmfamily\scriptsize,
    stage/.style={
        draw, rectangle, rounded corners=2pt,
        minimum width=2.45cm, minimum height=1.18cm,
        text width=2.22cm, align=center, inner sep=2pt,
        line width=0.5pt,
    },
    emphasis/.style={
        stage, line width=0.8pt, fill=gray!8,
    },
    shared/.style={
        draw, rectangle, rounded corners=2pt, dashed,
        minimum width=7.4cm, minimum height=0.82cm,
        text width=7.1cm, align=center, inner sep=2pt,
        line width=0.6pt,
    },
    flowarrow/.style={->, line width=0.5pt, draw=black},
    sharedlink/.style={line width=0.4pt, draw=black!65, dashed},
]

\node[anchor=west, font=\rmfamily\small\bfseries] at (0,4.08)
    {A\quad Offline teacher labelling and behaviour cloning};
\node[stage] (off-state) at (1.23,3.05)
    {State $s_t$\\+ up to $K$ targets};
\node[stage, right=0.30cm of off-state] (off-clone)
    {One cloned environment\\per candidate};
\node[stage, right=0.30cm of off-clone] (off-action)
    {Execute macro target $g$\\+ frozen S-tail};
\node[stage, right=0.30cm of off-action] (off-outcome)
    {Episode time and blood\\+ exact-safe status};
\node[emphasis, right=0.30cm of off-outcome] (off-label)
    {Teacher-selected\\candidate label};
\node[emphasis, right=0.30cm of off-label] (off-ranker)
    {Behaviour cloning\\$\rightarrow$ frozen ranker $\pi^{\mathrm R}$};

\draw[flowarrow] (off-state) -- (off-clone);
\draw[flowarrow] (off-clone) -- (off-action);
\draw[flowarrow] (off-action) -- (off-outcome);
\draw[flowarrow] (off-outcome) -- (off-label);
\draw[flowarrow] (off-label) -- (off-ranker);

\node[shared] (shared-rule) at ($(off-action.south)!0.5!(off-outcome.south)+(0,-0.88cm)$) {%
    \textbf{Shared frozen rollout rule:} simulator, S-tail, and budget predicate\\[-1pt]
    $B_{\mathrm{budget}}=B_{S,\mathrm{condition}}+M_B$; automatic clamp/unclamp
};
\draw[sharedlink] (off-action.south) -- (shared-rule.north west);
\draw[sharedlink] (off-outcome.south) -- (shared-rule.north east);

\node[anchor=west, font=\rmfamily\small\bfseries] at (0,0.62)
    {B\quad Online shielded execution};
\node[stage] (on-state) at (1.23,-0.42)
    {State $s_t$\\+ the same targets};
\node[emphasis, right=0.30cm of on-state] (on-ranker)
    {Deployed ranker $\pi^{\mathrm R}$\\orders candidates};
\node[stage, right=0.30cm of on-ranker] (on-shield)
    {Verify ranked prefix\\until pass or exhaustion};
\node[emphasis, right=0.30cm of on-shield] (on-admit)
    {First exact-safe target\\or fail closed};
\node[stage, right=0.30cm of on-admit] (on-transfer)
    {Shortest transfer\\+ cut/seal};
\node[stage, right=0.30cm of on-transfer] (on-next)
    {Next state\\$s_{t+1}$};

\draw[flowarrow] (on-state) -- (on-ranker);
\draw[flowarrow] (on-ranker) -- (on-shield);
\draw[flowarrow] (on-shield) -- (on-admit);
\draw[flowarrow] (on-admit) -- (on-transfer);
\draw[flowarrow] (on-transfer) -- (on-next);
\draw[sharedlink] (shared-rule.south) -- (on-shield.north);

\end{tikzpicture}
    \caption{Offline teacher labelling and online shielded execution. For each state, the generator returns up to $K$ candidates. Offline, one environment clone per candidate executes the macro target and frozen S-tail; the resulting full-episode time, simulated blood, and exact-safe status determine the teacher label used to train $\pi^{\mathrm{R}}$. Online, the frozen ranker orders candidates and the lazy exact shield checks them in that order, stopping as soon as one passes the same simulator, S-tail, and budget predicate. If no candidate passes, execution terminates as infeasible. ``Exact safety'' refers to this frozen simulator predicate, not to physiological or clinical safety.}
    \label{fig:learned-planner}
\end{figure*}
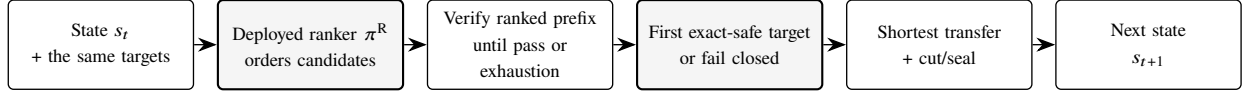

Figure~\ref{fig:learned-planner} summarizes how this shared candidate set feeds offline teacher labelling and online shielded execution.

\paragraph{Depth-one teacher and behaviour-cloned ranker.}
To label each candidate with complete-episode consequences, a depth-one teacher $\pi^{\mathrm{T}}_t$ executes $g\in\mathcal{G}_t$ on a cloned environment and then follows deterministic S-scan completion $\pi^{\mathrm{S}}$ until termination. This is equivalent in spirit to one-step tree search with a frozen completion policy. Each rollout reports
\begin{align}
T_{\mathrm{total}}(s_t,g) &= T_{\mathrm{past}}(s_t) + \Delta T(s_t,g) + T_{\mathrm{tail}}(s_t,g), \notag \\
B_{\mathrm{total}}(s_t,g) &= B_{\mathrm{past}}(s_t) + \Delta B(s_t,g) + B_{\mathrm{tail}}(s_t,g), \notag
\end{align}
Here, $\Delta T(s_t,g)$ and $\Delta B(s_t,g)$ are the incremental costs of executing candidate $g$ in the cloned environment, whereas the tail quantities are produced by completing the rollout with $\pi^{\mathrm{S}}$. To train only on admissible candidates, the teacher labels a rollout \emph{exact-safe} only when it completes without a failure reason and remains within $B_{S,\mathrm{condition}}+M_B$. It selects the accepted candidate that lexicographically minimizes $(T_{\mathrm{total}},B_{\mathrm{total}})$ with a quantised tie-break. If none is accepted, the teacher records \texttt{safety\_invariant\_violation} and excludes the state from behaviour-cloning labels. Exhaustively computing candidate outcomes is intrinsic to C3's minimization objective; it is not a separate deployment-shield cost.

To learn a compact approximation of the teacher's preference, the frozen ranker $\pi^{\mathrm{R}}$ is behaviour-cloned from 171,401 states and 1,017,114 candidate slots collected from 448 policy-training scenes under the default $15/5$ schedule and $p_{\mathrm{bleed}}=1.0$. The primary label is the teacher-selected candidate index. Multi-task heads also regress $(T_{\mathrm{total}},B_{\mathrm{tail}},B_{\mathrm{total}})$ for every valid candidate and classify \texttt{completion} and \texttt{safe\_exact}. The model uses a three-layer dilated convolutional encoder with $32$ channels and dilations $1,2,4$ over an $11$-channel semantic grid, a candidate MLP and global-context MLP with $96$ units each, a target-relative local feature, four masked mean/max pooled features, a $96$-unit trunk, and six output heads. Its fixed-weight objective is
\begin{align}
\mathcal{L} \;=\; &\,1.0\,\mathcal{L}_{\mathrm{rank}} + 0.2\,\mathcal{L}_{T_{\mathrm{total}}} + 0.3\,\mathcal{L}_{B_{\mathrm{tail}}} \notag \\
&+ 0.3\,\mathcal{L}_{B_{\mathrm{total}}} + 0.1\,\mathcal{L}_{\mathrm{completion}} + 0.2\,\mathcal{L}_{\mathrm{safe}}, \notag
\end{align}
where ranking uses cross-entropy over the state's accepted set, numeric heads use masked MSE, and binary heads use sigmoid cross-entropy. Optimisation uses AdamW with learning rate $3\times10^{-4}$, weight decay $10^{-5}$, gradient clipping at $5.0$, batch size $256$, and five epochs. The released checkpoint was trained with seed $2026081603$. To support reproduction, the released artifacts bundle the encoder, loss weights, and feature scales; \ref{app:bc-config} and Table~\ref{tab:bc-config} define the fixed values and their meanings.

\paragraph{Lazy exact simulator-rollout shield and computational note.}
To separate learned preference from execution permission, C4 first uses $\pi^{\mathrm{R}}$ to produce the deterministic order $(g_{(1)},\ldots,g_{(K)})$ of $\mathcal{G}_t$. The lazy exact shield then evaluates candidates in that order. Each check executes $g_{(j)}$ on a cloned environment and runs the same frozen $\pi^{\mathrm{S}}$ tail used by the teacher \cite{Alshiekh2018Shielding}. Define
\begin{equation}
\begin{aligned}
E_t(g)=\mathbf{1}\bigl[{}&\mathrm{complete}(s_t,g)\ \land\ f(s_t,g)=\varnothing\\
&\land\ B_{\mathrm{total}}(s_t,g)\le B_{S,\mathrm{condition}}+M_B\bigr],
\end{aligned}
\end{equation}
where $f(s_t,g)$ is the rollout failure reason. The per-scene baseline $B_{S,\mathrm{condition}}$ is the deterministic serpentine completion's total simulated blood under the same clamp and $p_{\mathrm{bleed}}$ condition. The fixed margin $M_B=16.071$ mL equals $5\%$ of mean C0 blood on the policy-training split and is reused for evaluation. C4 executes
\begin{equation}
j^*=\min\{j\in\{1,\ldots,K\}:E_t(g_{(j)})=1\}
\end{equation}
and stops verification immediately when $j^*$ is found. If the set is empty, the controller terminates with \texttt{infeasible\_no\_safe\_candidate}; it does not execute an unverified or rejected fallback. The term \emph{shield} denotes this policy-external software predicate and does not imply physiological or clinical safety.

Because $j^*$ is exactly the highest-ranked admissible candidate, lazy stopping preserves the selection defined by the complete admissible set while evaluating only the prefix required to find it. Permission remains independent of learned scores because $E_t$ uses the frozen simulator and frozen $\pi^{\mathrm{S}}$ tail rather than the ranker's safety head. A 256-scene audit produced identical action-sequence hashes to complete admissible-set selection. In the main confirmation set, C4 evaluated 97,421 exact rollouts for 97,320 macro actions (1.001 per action): the first-ranked candidate was accepted in 97,228 actions, the second in 83, and the third in nine. By contrast, C3 must compute all candidate rollouts to identify its exact full-episode minimum. This difference is the source of C4's reduced planning latency.

\paragraph{Why most C4 macro actions follow the serpentine and what is not learned.}
To preserve the strong deterministic baseline, the teacher selects a non-serpentine candidate only if it passes the exact safety predicate and improves the lexicographic pair of full-episode time and total blood, with fixed tie-breaking. In most states, the S target is already accepted and no alternative is preferable under this rule. Consistent with this behavior, the frozen C4 checkpoint selected the S target for $98.97\%$ of macro actions on the 256-scene evaluation split, with state-dependent deviations in the remaining $1.03\%$. Controller C0 in Table~\ref{tab:controllers}, which removes learned ranking, recovers the deterministic S baseline. Controller C5, which removes the shield, reintroduces rare budget overruns. In summary, the model determines candidate preference, the lazy exact shield determines execution eligibility, and deterministic shortest-path transfer performs low-level routing; none of these components establishes physiological validity.

\FloatBarrier

\section{System Demonstration and Functional Verification}

This chapter evaluates VoxelSage at three complementary levels. Section~\ref{sec:online-portb-verification} verifies Port~B coordinate handling, skill integration, three-dimensional-to-planar transfer, and fail-closed behaviour through synthetic phantoms, public CT cases, and constructed surfaces. Section~\ref{sec:frozen-confirmation} measures the learned controller against frozen comparators in the independent planar simulator, including shield ablation, latency, and condition sensitivity. Section~\ref{sec:porta-evaluation} evaluates whether Port~A can interpret requests, invoke the appropriate skills, preserve evidence faithfulness, manage multiple cases, and reuse cached results. Because these streams test different claims, their results are interpreted separately.

\subsection{Online System Demonstration and Port B Verification}\label{sec:online-portb-verification}

The online evaluation proceeds through four stages:
\begin{enumerate}
    \item verification of physical-coordinate correctness using synthetic NIfTI phantoms; 
    \item case preparation and execution of all eight built-in Port~B skills on three public cases;
    \item learned-controller transfer between a confirmed surface and the frozen planar canvas (3D $\rightarrow$ 2D);
    \item evaluation of fail-closed and state-invalidation behavior.
\end{enumerate}
This organization separates numerical correctness, workflow coverage, learned bridging, and defensive behavior instead of treating them as one undifferentiated demonstration.

The physical-coordinate stage uses only synthetic non-medical data. The case-based functional and defensive checks use NIfTI conversions and released reference masks from the TCIA \emph{Colorectal-Liver-Metastases} Version~2 collection, which contains preoperative CT, DICOM segmentations, and clinical data for 197 subjects under CC BY~4.0 \cite{Simpson2023CRLMDataset,Clark2013TCIA}.
Reference masks isolate the downstream integration tests from segmentation error: variability in VISTA3D inference could otherwise make a failed measurement, reconstruction, or planning path ambiguous. VISTA3D's segmentation performance and automatic and prompt-guided operating modes were evaluated in its original study \cite{He2025VISTA3D}; because the present contribution is system integration rather than a segmentation model, we do not repeat that model-level validation. 
The three prespecified public cases used for the evaluation are \texttt{CRLM-CT-1007}, \texttt{CRLM-CT-1012}, and \texttt{CRLM-CT-1036}.

\subsubsection{Stage 1: Physical-Coordinate Correctness}

The test set comprises 24 synthetic NIfTI phantoms spanning isotropic, anisotropic, and rigidly rotated affine families, together with large and small components, axis-separated and overlapping masks, empty masks, and below-threshold inputs. Of the 216 direct-function and skill-API checks, 186 were numerical comparisons and 30 tested structured empty or below-threshold behaviour. All numerical comparisons passed the absolute tolerance of 0.01 in the corresponding metric unit, and all boundary-condition (empty and invalid conditions) checks returned the expected structured result. The anisotropic and rotated families agreed with affine-derived ground truth. These results verify coordinate and unit handling for the constructed phantoms rather than measurement accuracy on clinical images.

\subsubsection{Stage 2: Three-Case Port B Functional Matrix}

\begin{table*}[t]
\centering
\footnotesize
\caption{Port~B skill verification on three public case aliases. The five measurement/slice skills summarize nine calls each (three per case); reconstruction, editing, and sequence times use one initial reconstruction, saved edit, and nearest traversal per case, respectively. Times are descriptive shared-server measurements.}
\label{tab:portb-skill-matrix}
\renewcommand{\arraystretch}{1.08}
\begin{tabularx}{\textwidth}{@{}>{\raggedright\arraybackslash}p{0.19\textwidth}cX>{\raggedleft\arraybackslash}p{0.19\textwidth}@{}}
\toprule
Skill & Cases passed & Required output or state verified & Median s (range) \\
\midrule
Liver analysis & 3/3 & Structured volume, diameter, distance, and report & 5.906 (5.441--6.468) \\
Tumour diameter & 3/3 & Per-lesion diameter, method, and voxel count & 0.197 (0.180--0.305) \\
Tumour--vessel distance & 3/3 & Hepatic and portal distance, contact, and mask variant & 5.913 (5.621--24.756) \\
Vessel volume & 3/3 & Volume in mm$^3$/cm$^3$, voxel count, and mask variant & 0.202 (0.193--0.388) \\
Slice selection & 3/3 & Three raw/overlay pairs with artifact hashes & 3.509 (3.153--24.126) \\
3D reconstruction & 3/3 & HTML, scene JSON, anatomy list, and candidate surface & 33.368 (29.835--74.109) \\
Segmentation editing & 3/3 & Edited-mask hash, backup, and stale-scene invalidation & 3.578 (2.763--4.743) \\
Resection sequence & 3/3 & Saved surface, legal-start preview, and nearest traversal & 23.920 (5.047--68.636) \\
\bottomrule
\end{tabularx}
\end{table*}

On each frozen public case, we tested cold and cached case preparation, repeated read-only skill calls, reconstruction before and after a saved mask edit, segmentation editing, and resection-sequence preview or execution. Cold preparation took a median 43.3~s (range 43.1--65.0~s); cached reuse was reported as 0.0~s at the service's one-decimal precision, with unchanged CT and mask hashes. All 66 recorded checks passed, covering all 24 unique skill--case paths (eight skills across three cases; Table~\ref{tab:portb-skill-matrix}).

\subsubsection{Stage 3: Learned-Controller Transfer}

The current core-only adapter was checked on constructed surfaces with no vessel, an internal vessel, a boundary-vessel barrier, and disconnected target regions. On identical inputs, lazy and complete candidate verification produced identical macro-action hashes, replayed paths, coverage, simulated time and blood, budgets, and intervention counts. Three additional constructed surfaces were each run twice through the public skill entry point, checking dynamic loading, saved-surface input, vessel retention, complete-grid mapping, and deterministic output.

The archived three-case bridge used support-expanded targets and is not counted as validation of the current core-only adapter. Its retained artifacts do not identify all core/support cells needed for an exact replay; validation on patient-derived core-only surfaces therefore remains outstanding. The constructed checks establish implementation compatibility only. Controller time and simulated-blood effects are evaluated separately on the independent 256-scene planar set in Section~\ref{sec:frozen-confirmation}.

\subsubsection{Stage 4: Fail-Closed and State-Invalidation Behavior}

Nine prespecified defensive checks all met their expected outcomes. Missing or invalidated surfaces, missing or hash-mismatched checkpoints, non-frozen cell size, incompatible grid override, and illegal start cell produced visible structured rejection. When we deliberately disconnected the surface, the planner marked the result as \texttt{partial} instead of incorrectly reporting complete coverage. Separately, referring to the same checkpoint file through a different path did not change its identity: the system reported the same frozen hash and learned policy identifier.

The sealed experimental bundle matched every recorded SHA-256 entry, supporting artifact traceability rather than software correctness or patient-level validity.

\subsection{Planar-Simulator Evaluation}\label{sec:frozen-confirmation}

\begin{figure*}[t]
\centering
\includegraphics[width=0.9\textwidth]{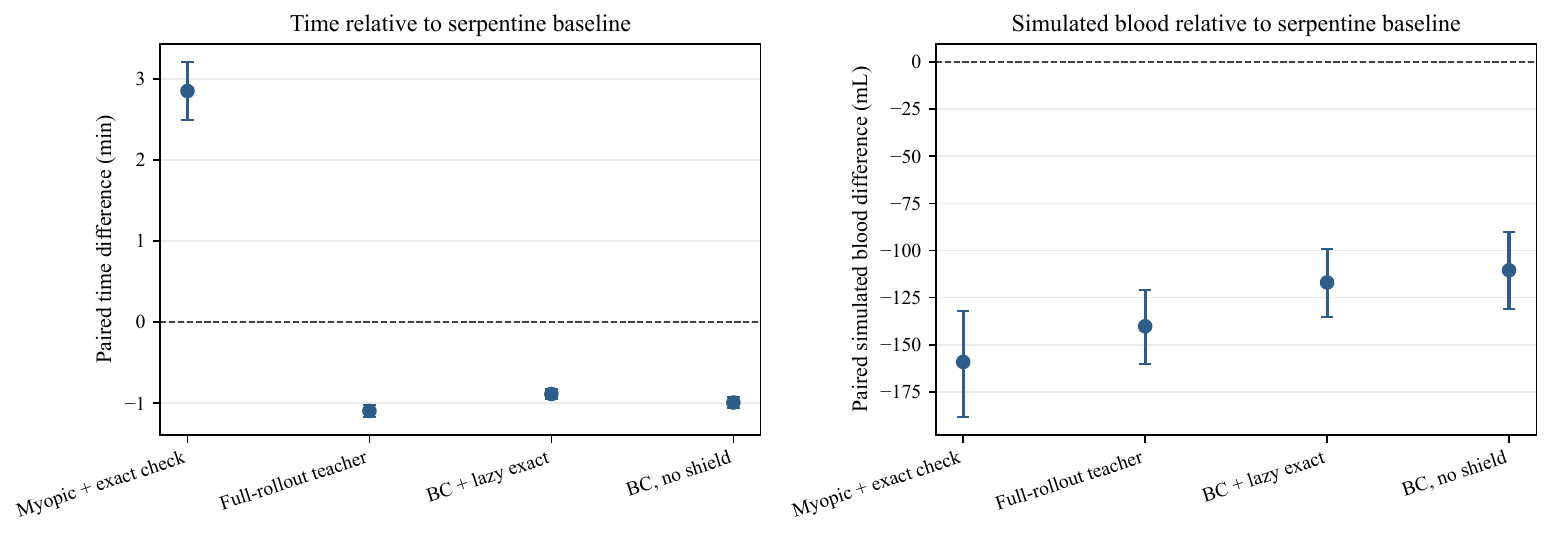}
\caption{Controller effects on the 256-scene post-freeze replication set relative to the deterministic serpentine baseline. Points show paired means and whiskers show 95\% paired-bootstrap confidence intervals. Simulated blood is an uncalibrated environment quantity.}
\label{fig:replication-effects}
\end{figure*}

\begin{figure}[t]
\centering
\includegraphics[width=\linewidth]{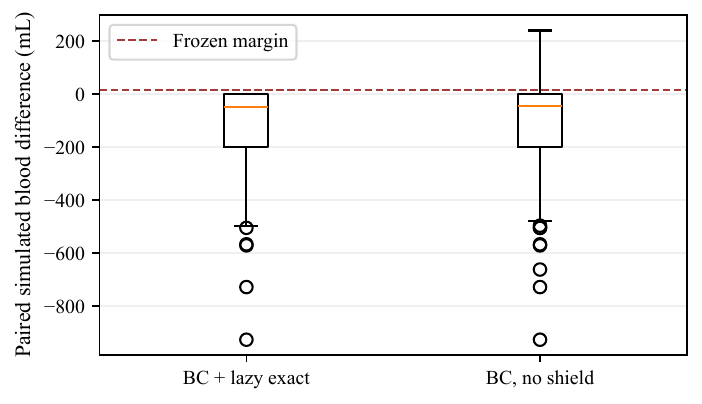}
\caption{Shield ablation on the 256-scene v10.8 confirmation set. Boxplots show paired simulated-blood differences from C0 for shielded C4 and unshielded C5; the dashed line marks the frozen budget margin above the C0 baseline.}
\label{fig:shield-ablation}
\end{figure}

\subsubsection{Protocol and controller definitions}

The planar study uses synthetic two-dimensional resection-planning scenes independent of the CT demonstration. Disjoint policy-training (448), internal-development (64), tuning (64), validation (128), one-shot test (128), and stress (128) splits were frozen. Policy training alone supplied labels, gradients, and normalization scales; internal development supported teacher admission, early stopping, and the simulator-level development gate; tuning compared prespecified behaviour-cloning configurations; and validation selected the final configuration, training seed, and checkpoint. The one-shot test and subsequent stress split provided, respectively, locked in-distribution evaluation and an out-of-distribution audit, and neither informed model selection. The released five-epoch checkpoint, trained with seed~2026081603, was selected under the prespecified validation rule. Appendix~\ref{app:bc-config} details the split roles and development gate.

After checkpoint freeze, we used 32 scenes for a pilot, a newly frozen set of 256 previously unseen scenes for the main confirmation, 64 of those confirmation scenes for an isolated latency benchmark with three repetitions, and 128 base geometries per sensitivity condition. Table~\ref{tab:controllers} defines the controller roles; the main comparison reports C0, C2, C3, C4, and C5. The primary endpoint is the paired time difference C4--C0; C4--C2 tests learning specificity, C3 is the full-rollout teacher reference, and C5 isolates removal of exact verification. Confidence intervals use 10,000 paired bootstrap resamples. The simulator gate requires a complete legal episode, no invariant or budget violation, a paired simulated-blood increase within $M_B$, and a favorable time interval excluding zero.

The sensitivity audit uses a condition-specific episode budget for each clamp and bleeding condition while keeping the checkpoint and $M_B$ fixed. Completion, infeasibility, and realized-budget overrun are reported separately; efficacy differences are not interpreted for conditions in which C4 does not complete every scene.

The reported experiments ran locally on a Windows~11 workstation with an Intel Core Ultra~7 255HX CPU (20 logical processors), Python~3.10.21, and PyTorch~2.5.1+cpu. The isolated latency benchmark used one scene worker, three repetitions per scene, and six leaf workers for C3; its wall times must not be pooled with results from the earlier server environment or with multi-scene throughput runs.

\subsubsection{Primary replication and learning specificity}

All five reported controllers completed 256/256 main-confirmation scenes with legal actions. C4 reduced mean simulated time from 34.274 to 33.388 min relative to C0, a paired difference of $-0.886$ min (95\% CI $[-0.947,-0.826]$; Cohen's $d_z=-1.77$), with 252 wins, three ties, and one loss. Its mean simulated-blood difference was $-116.995$ mL (95\% CI $[-135.484,-99.784]$); C4 never exceeded C0 beyond numerical tolerance and produced no episode-budget overrun.

C4 was 3.736 min faster than the non-learned C2 comparator (95\% CI $[-4.094,-3.389]$; $d_z=-1.28$), supporting learning specificity against that heuristic rather than superiority over every deterministic method. The full-rollout teacher C3 achieved a slightly shorter simulated episode time: 33.178 min versus 33.388 min for C4, a mean C4--C3 difference of $+0.210$ min. C4 nevertheless retained 80.9\% of the teacher's improvement over C0 while avoiding C3's need to evaluate all candidates. Figure~\ref{fig:replication-effects} summarizes controller-level paired effects on simulated time and blood relative to C0.

\subsubsection{Shield ablation and computational cost}

The unshielded C5 controller exceeded its episode budget in 24/256 scenes, compared with 0/256 for C4. The lazy exact shield changed the top-ranked choice in only 92/97,320 macro actions (0.0945\%), yet those sparse interventions removed the observed nominal-set overruns. Figure~\ref{fig:shield-ablation} shows the paired simulated-blood differences from C0; values above the frozen margin correspond to episode-budget overruns.

In the isolated local CPU benchmark (64 scenes, three repetitions), each scene was first summarized by its median across repetitions. Across these 64 scene medians, C4 had median, mean, and 95th-percentile episode wall times of 21.21, 27.69, and 65.10~s. C3, with six leaf workers, required 46.85, 74.43, and 267.02~s, respectively. Thus, C4's median planning time was 54.7\% lower, or approximately $2.21\times$ faster, than C3. This comparison measures controller computation rather than simulated surgical time: C3 still produced episodes that were 0.210 simulated minutes shorter on average, whereas C4 provided the intended latency--quality compromise. C5 omits exact verification and therefore represents the less constrained ablation, not the primary controller.

\subsubsection{Condition-specific sensitivity}\label{sec:sensitivity}

C4 completed all 128 scenes under each of S0--S4 with no infeasible termination or episode-budget overrun. C5 also completed every scene, but exceeded its condition-specific budget in 14, 10, 12, 11, and 8 scenes, respectively (Table~\ref{tab:corrected-sensitivity}). These results support completion and budget adherence under the tested perturbations. The lower bleeding coefficients retain the same absolute margin $M_B$, so their relative budget allowance can be larger; they do not establish robustness under uniformly scaled relative constraints.

\begin{table}[t]
\centering
\small
\caption{Sensitivity audit. C4 fail denotes termination without complete coverage; overrun denotes realized episode blood above the condition-specific budget. C5 completed all scenes.}
\label{tab:corrected-sensitivity}
\begin{tabular}{@{}lcc@{}}
\toprule
Condition & C4 complete/fail & Overrun C4/C5 \\
\midrule
S0: 15/5, $p=1.00$ & 128/0 & 0/14 \\
S1: 12/5, $p=1.00$ & 128/0 & 0/10 \\
S2: 10/5, $p=1.00$ & 128/0 & 0/12 \\
S3: 15/5, $p=0.50$ & 128/0 & 0/11 \\
S4: 15/5, $p=0.25$ & 128/0 & 0/8 \\
\bottomrule
\end{tabular}
\end{table}

The main confirmation, learning-specificity, shield-ablation, and latency results form an internally consistent evidence chain inside the frozen simulator. The sensitivity audit confirms complete coverage without observed budget overruns across the five tested conditions using condition-specific baselines. None of these experiments validates the online three-dimensional adapter, a clinical workflow, a surgical trajectory, physiological safety, or patient-level blood loss; those distinctions motivate the separate Port~B bridge and fail-closed experiments above.

\subsection{Port A Orchestration Evaluation}
\label{sec:porta-evaluation}

\paragraph{Evaluation objectives}
While the independent evaluations of Port B in Sections~\ref{sec:online-portb-verification}
and~\ref{sec:frozen-confirmation} establish its correctness within the tested scope,
correct computation by Port B alone does not ensure users can get correct answer after entering queries, which requires the correctness of the whole workflow of our website, including both skill execution and agent orchestration. We therefore evaluate the ability of Port~A, determining whether Port A can correctly interpret requests, invoke the
appropriate skills, and formulate answers from their outputs.
To assess the reliability and efficiency of this process,
we also examine whether Port A avoids unsupported findings (i.e. hallucination), keeps information associated with the correct case, and
correctly reuses available results.

\paragraph{Data and experimental tasks}
We designed four experiments: Task Execution, Multi-Case
Management, Evidence Faithfulness, and Cache Efficiency.
We selected 63 cases using CT scans from
AbdomenAtlas1.0Mini \cite{abdomenatlas_ct} with reference masks from
AbdomenAtlas3.0Mini \cite{abdomenatlas_masks}, and
DeepTumorVQA (a three-dimensional CT question-answering
benchmark covering recognition, measurement, visual reasoning,
and medical reasoning \cite{chen2026deeptumorvqa}) questions supported by the available skills.
To isolate measurement and orchestration from segmentation
errors, all image-based measurement experiments used
reference masks instead of automatic segmentation outputs.

Task Execution comprised 154 tasks: 44 lesion-existence,
34 lesion-counting, 24 liver-volume, and 27 total-lesion-volume
questions retained their original DeepTumorVQA wording and
answers. For the 25 diameter tasks, we explicitly asked for the
maximum three-dimensional lesion diameter and used
PyRadiomics \cite{vanGriethuysen2017} to compute the corresponding reference answers,
because we could not verify which diameter definition
was used in the original DeepTumorVQA question--answer pairs.

Multi-Case Management comprised 63 paired-case liver-volume
comparisons. Evidence Faithfulness used 137 questions with
case IDs that did not exist in Port B.
Cache Efficiency repeated 137 questions in their original
sessions to assess result reuse.

\paragraph{Execution settings}
To test actual tool execution, we used Qwen3.8-Flash in
Port A and invoked skills through Port B's HTTP API.
Each request allowed up to four agent rounds and
180 seconds, with output limits of 700 tokens for the
original-question tests and 400 tokens for the diameter tests.
Task Execution, Multi-Case Management, and Evidence
Faithfulness used fresh sessions without cached measurements
so that answers depended on evidence obtained during the
current task. Cache Efficiency retained the initial tool
results and repeated the question in the same session
to measure the benefit of reuse.

\begin{table*}[t]
    \centering
    \caption{Results for the four
    Port A experiments.}
    \label{tab:porta-results}
    \small
    \renewcommand{\arraystretch}{1.15}
    \begin{tabularx}{\textwidth}{@{}
        p{0.22\textwidth}
        X
        p{0.23\textwidth}@{}}
        \toprule
        Experiment & Metric & Result \\
        \midrule

        Task Execution &
        Lesion-existence accuracy &
        44/44 (100\%) \\

        & Lesion-counting accuracy &
        30/34 (88.2\%) \\

        & Liver-volume accuracy &
        22/24 (91.7\%) \\

        & Total-lesion-volume accuracy &
        27/27 (100\%) \\

        & Maximum three-dimensional
        lesion-diameter accuracy &
        25/25 (100\%) \\
        \midrule

        Multi-Case Management &
        Correct identification of the case
        with the larger liver volume &
        59/63 (93.7\%) \\
        \midrule

        Evidence Faithfulness &
        Responses without invented measurements &
        137/137 (100\%) \\
        \midrule

        Cache Efficiency &
        Repeated queries without additional tool calls &
        130/137 (94.9\%) \\

        & Median response time:
        initial $\rightarrow$ repeated query &
        8.16 $\rightarrow$ 2.56~s \\

        & Reduction in median response time &
        68.6\% \\

        & Answer accuracy under the batch's
        original scoring criteria &
        122/137 (89.1\%) \\
        \bottomrule
    \end{tabularx}

    \vspace{4pt}
    \begin{minipage}{0.98\textwidth}
        \footnotesize
        \textit{Note.}
        The 27 total-lesion-volume questions represent
        15 distinct cases. The total-lesion-volume and diameter
        results were obtained in separate evaluations.
        The Cache Efficiency experiment used the original VQA
        diameter references and did not include the lesion-volume
        evaluation reported under Task Execution, so its accuracy
        is not directly comparable with the Task Execution results.
    \end{minipage}
\end{table*}

\paragraph{Scoring criteria}
We matched the metrics to the purpose of each experiment:
\begin{itemize}
    \item \textbf{Task Execution}
    Accuracy was the proportion of final answers meeting
    the reference criteria; unanswered tasks counted as
    incorrect. Lesion existence and counts required exact
    agreement because their answers are categorical or
    integer-valued. For volume measurements, we combined
    relative tolerances with absolute lower limits to
    accommodate differences in size and small numerical
    deviations. The allowed absolute error was
    $\max(1~\mathrm{mL},\,0.02V_{\mathrm{ref}})$
    for liver volume and
    $\max(0.1~\mathrm{mL},\,0.05V_{\mathrm{ref}})$
    for total lesion volume, where $V_{\mathrm{ref}}$
    is the reference volume in mL.
    These were experimental agreement thresholds rather
    than clinical error standards.
    Diameter required agreement within 0.01~mm of the
    PyRadiomics reference to allow numerical and rounding
    differences under the same measurement definition.

    \item \textbf{Multi-Case Management}
    Accuracy was the proportion of comparisons that
    correctly identified the case with the larger
    liver volume.

    \item \textbf{Evidence Faithfulness}
    The primary metric was the proportion of responses
    without invented measurements when the case was missing.
    A correct response should state that data were
    unavailable and return \texttt{null} in the structured
    numerical field.

    \item \textbf{Cache Efficiency}
    We measured the proportion of repeated queries without
    additional tool calls and compared median response
    times before and after reuse. Answer accuracy was
    also reported to assess whether faster responses
    remained correct.
\end{itemize}

\paragraph{Results}
Table~\ref{tab:porta-results} reports the results for each
experiment. Total-lesion-volume and diameter results were
obtained in separate evaluations, while Cache Efficiency
retains the original VQA scoring criteria.
We therefore report the results separately without
a pooled overall accuracy.

The results support the four abilities to different degrees:
\begin{itemize}
    \item \textbf{Task Execution}
    Perfect scores for lesion existence, total lesion volume,
    and maximum three-dimensional lesion diameter indicate
    good performance under the tested conditions.
    Errors in lesion counting and liver volume show
    remaining limits in these measurement tasks.

    \item \textbf{Multi-Case Management}
    Most comparisons were correct, supporting the ability
    to associate measurements with the right cases,
    although correct comparison was not guaranteed.

    \item \textbf{Evidence Faithfulness}
    No measurements were invented when cases were missing,
    supporting appropriate handling of unavailable
    evidence in this setting.

    \item \textbf{Cache Efficiency}
    Fewer tool calls and shorter response times show
    the benefit of result reuse, but reuse alone
    does not ensure correct answers.
\end{itemize}

\paragraph{Scope and limitations}
Reference masks excluded automatic segmentation errors,
so the results do not evaluate performance with predicted
masks. Evidence Faithfulness tested only nonexistent cases,
not existing cases with missing masks or incomplete tool
results. Those scenarios require Port A to distinguish
available evidence from missing evidence; the observed
absence of invented measurements therefore cannot be
extended to all evidence failures.
Finally, the accuracy thresholds measure agreement with
reference answers or calculations under experimental
criteria rather than clinically acceptable error.
Whether these results meet the requirements of clinical
use remains to be assessed through further evaluation.

\section{Conclusion}
VoxelSage connects case-centered CT preparation, deterministic measurement, selected two- and three-dimensional evidence, interactive correction, and confirmation-gated planning through a dual-port agent architecture. Results remain linked to source images, masks, geometry, and execution records. The optional planning adapter uses a geometry-only conservative-core crop that retains boundary vessel proxies to map a frozen planar ranker and lazy exact simulator shield onto a confirmed surface while preserving complete-grid identifiers and fail-closed behavior. Constructed-surface tests verified the adapter and its public skill invocation; in the separate frozen simulator C4 improved time over prespecified baselines without nominal-set budget violations while reducing median controller wall time by 54.7\% relative to C3. The same frozen controller completed all five sensitivity conditions without observed episode-budget overruns.

These results establish workflow integration and a reproducible simulator experiment, rather than clinical accuracy, benefit, safety, or patient-level blood-loss prediction. The system has been evaluated only as an in-lab engineering demonstration on a small number of public cases, and its learned component was trained on two-dimensional proxy tasks rather than real surgical dynamics. Clinical utility will require external evaluation in representative cohorts and real clinical workflows, with appropriate outcome measures and uncertainty estimation.

\raggedbottom
\bibliography{references}

\flushbottom
\clearpage
\appendix
\renewcommand{\thesection}{Appendix \Alph{section}}
\renewcommand{\thesubsection}{\Alph{section}.\arabic{subsection}}
\renewcommand{\thesubsubsection}{\Alph{section}.\arabic{subsection}.\arabic{subsubsection}}
\section{Skill Definitions and Tool-Calling Examples}
\label{app:skills}

This appendix documents the built-in skills used in the engineering evaluation and provides representative interface examples. It complements the Medical Imaging Skills section (Section~\nameref{sec:imaging-skills}) in the main text by making the capability boundary, manifest contract, and returned evidence concrete; it does not introduce additional evaluation claims.

\subsection{Available Skills}

Table~\ref{tab:builtin_skills} summarizes the eight built-in Port~B skills, grouped into four categories as in Section~\nameref{sec:imaging-skills}: quantitative measurement skills return deterministic mask-derived quantities, evidence-generation skills create inspectable two- or three-dimensional artifacts, the interactive correction skill revises masks under explicit user control, and the planning skill operates only on a user-saved candidate surface. The corresponding \texttt{skill.yaml} manifests and \texttt{main.py} entry points are available at \url{https://github.com/ZJUMAI/VoxelSage/tree/main/Port_B/skills/builtin}. The registry may additionally contain user-provided skills, which are not included in the reported evaluation.

\begin{table*}[t]
\centering
\small
\caption{Built-in Port~B skills in VoxelSage.}
\label{tab:builtin_skills}
\renewcommand{\arraystretch}{1.12}
\begin{tabularx}{\textwidth}{@{}p{0.23\textwidth}p{0.27\textwidth}X@{}}
\toprule
Skill & Primary function & Main output or prerequisite \\
\midrule
\texttt{liver\_analysis} & Integrated liver, vessel, and lesion analysis & Liver and vessel volumes, lesion diameters, tumor--vessel distances, and formatted report \\
\texttt{tumor\_diameter} & Physical-space maximum lesion diameter & Per-lesion diameter, method, and voxel count \\
\texttt{tumor\_vessel\_distance} & Tumor proximity to portal and hepatic veins & Per-lesion distance, contact flag, and vessel label \\
\texttt{vessel\_volume} & Portal- and hepatic-vein volume measurement & Volume in $\mathrm{mm}^3$ and $\mathrm{cm}^3$ \\
\texttt{slice\_selection} & Informative axial-slice ranking and export & Ranked raw CT and overlay PNG pairs \\
\texttt{three\_d\_reconstruction} & Mesh reconstruction and optional resection-surface generation & Three.js view, scene JSON, and optional B\'ezier candidates \\
\texttt{segmentation\_\allowbreak modification} & User-initiated mask-refinement session & Editable mask session and revised NIfTI mask after saving \\
\texttt{plan\_resection\_sequence} & Deterministic surface coverage or explicit experimental \texttt{learned\_shielded} ordering & Path JSON; source/adapter grids, crop/boundary-vessel and provenance metadata; or fail-closed prerequisite/checkpoint error \\
\bottomrule
\end{tabularx}
\end{table*}

Collectively, these skills separate case preparation and deterministic computation from orchestration and review. The measurement and evidence-generation skills are read-only and stateless, whereas the correction and planning skills introduce explicit user state---a saved mask or a saved candidate surface---that gates downstream recomputation and planning. Their shared manifest and case-context contract allows Port~A to select among them without exposing unrestricted case paths or treating every capability as one monolithic pipeline.

\subsection{Manifest, User-Skill Package, and Registration Contract}\label{sec:user-skill-contract}

Each built-in skill is defined by a \texttt{skill.yaml} manifest that records its \texttt{name}, \texttt{version}, \texttt{description}, activation \texttt{triggers}, and JSON-style \texttt{inputs} and \texttt{outputs}. Context-owned inputs, including the CT path and mask directory, are resolved by Port~B from the case identifier rather than supplied as unrestricted paths by the agent. Listing~\ref{lst:manifest-schema} shows an abbreviated version of the integrated liver-analysis manifest; the complete manifests preserve additional descriptions, nested measurement fields, and constraints.

\begin{lstlisting}[
caption={Abbreviated manifest schema for the integrated liver-analysis skill.},
label={lst:manifest-schema}]
name: liver_analysis
version: 1.0.0
inputs:
  ct_nifti_path:
    type: string
    source: context
    required: true
  mask_dir:
    type: string
    source: context
    required: true
outputs:
  type: object
  properties:
    liver_volume_cm3: {type: number}
    vessel_volumes: {type: object}
    tumor_results: {type: object}
    report_text: {type: string}
\end{lstlisting}

Custom skills reuse the same manifest model. A user-provided package must satisfy the following requirements:

\begin{enumerate}
    \item \textbf{Package structure.} A minimal package directory contains \texttt{skill.yaml} and \texttt{main.py}. The repository includes \texttt{examples/skills/tumor\_volume\_report} as a minimal example.
    \item \textbf{Manifest schema.} The manifest must define \texttt{name}, \texttt{version}, and \texttt{description}; \texttt{triggers}, \texttt{inputs}, and \texttt{outputs} are optional. Inputs marked \texttt{source: context} are excluded from the language-model function schema and supplied by Port~B, whereas other inputs become task parameters exposed to the model.
    \item \textbf{Entry point.} The entry point must define \texttt{run(ctx)} and return a JSON-serializable dictionary. Its context exposes the case identifier and parameters together with lazy accessors for the CT array, affine, voxel spacing, available masks, mask arrays or paths, output directory, and logging.
    \item \textbf{Registration.} The frontend skill manager accepts one \texttt{.zip} package, refreshes the skill list after registration, labels entries as built-in or user supplied, and exposes deletion for user entries. The registration engine requires both package files, restricts the manifest name to letters, digits, hyphens, or underscores, searches \texttt{main.py} for a \texttt{run} definition, limits each file to 2~MiB and the directory to 5~MiB, and copies the directory to \texttt{skills/user/\{name\}}. Registering the same name replaces that directory and its in-memory entry; unregistering a user skill removes the copy.
    \item \textbf{Execution and isolation.} \texttt{GET /api/skills/list} exposes registered manifests in both human-readable and OpenAI-compatible function-calling forms, and \texttt{POST /api/skills/run} supplies the prepared case context. Built-in skills are imported into Port~B, whereas a user skill runs in a child Python process. 
    The wrapper enforces a 120-s wall timeout and operating-system limits of 120 CPU seconds, 32 processes, and 512 open files, then accepts only a dictionary that can be serialized as JSON. These controls bound some failures but do not provide a separate operating-system identity, container, memory quota, dependency environment, or filesystem and network isolation; the child receives paths to case inputs and the output directory.
\end{enumerate}

At the invocation level, a client prepares a case through \texttt{POST /api/process-lite}, discovers skills through \texttt{GET /api/skills/list}, and invokes a selected skill with its \texttt{case\_id} and optional \texttt{params} through \texttt{POST /api/skills/run}. This contract keeps case-owned artifacts inside the medical-computing service while exposing task-specific parameters and structured results to the orchestrator.

\subsection{Tool-Calling Examples}

The following abbreviated API invocation uses illustrative case \texttt{CRLM-CT-1012} and is separate from the controlled evaluation.

\begin{lstlisting}[
caption={Request for integrated liver analysis.},
label={lst:liver-analysis-request}]
POST /api/skills/run
{
  "skill_name": "liver_analysis",
  "case_id": "CRLM-CT-1012",
  "params": {}
}
\end{lstlisting}

Listing~\ref{lst:liver-analysis-response} shows the common response envelope and selected result fields.

\begin{lstlisting}[
caption={Abbreviated structured response from the liver-analysis skill.},
label={lst:liver-analysis-response}]
{
  "status": "ok",
  "result": {
    "liver_volume_cm3": 871.76,
    "tumor_results": {
      "tumor_1": {
        "diameter": {"max_diameter_mm": 16.32},
        "vessel_distances": {
          "hepatic": {"min_distance_mm": 35.04}
        }
      }
    }
  },
  "execution_time_ms": 17828.0
}
\end{lstlisting}

The complete result also contains voxel counts, methods, contact flags, mask variants, and a formatted report.

\subsection{Experimental Learned-Planning Contract}

The \texttt{learned\_shielded} algorithm is opt-in and does not replace the deterministic \texttt{nearest}, \texttt{dfs}, or \texttt{spanning\_tree} modes. It requires an explicitly saved surface, approximately 4-mm cells whose cropped liver-domain window fits the frozen $30\times40$ canvas, and checkpoint SHA-256 \breakablett{c07904502d6b71a74484adb1c27971c77cdf6a61bb20b04f1f39d786d61a70be}. With the default intersection threshold, the adapter retains the four-of-five sampled-liver core without adding support cells, preserves boundary vessel proxies in place, and maps the cropped adapter indices back to the complete saved-surface grid. Boundary components follow the full-ring rule except for the frontier-deadlock release described in Section~\ref{sec:sequential-planning}. At each macro step, the frozen ranker orders the candidates and the lazy exact shield verifies them in that order, stopping at the first candidate that completes under the frozen S-tail without failure or budget excess.

A missing or mismatched checkpoint, incompatible cell size, out-of-envelope cropped domain, illegal start, incomplete controller rollout, absence of an admissible candidate, or simulator-budget overrun produces a structured failure rather than a deterministic fallback labelled as learned output. The controller covers the start-connected component; when other target components remain, the skill returns \texttt{partial}, identifies uncovered cells, and reports coverage against the complete target. Successful results distinguish three-dimensional geometry from simulator proxies by recording source and adapter grids, crop origin, core and boundary-vessel counts, mapped three-dimensional path, policy and checkpoint identifiers, intervention count, simulator time and blood quantities, runtime, and a scope warning. These fields support implementation audit, not clinical validation of the path, vessel proxy, or simulator budget.

\clearpage
\onecolumn
\section{Behaviour-Cloning Reproduction Configuration}\label{app:bc-config}

\subsection{Frozen data separation and development gate}

Table~\ref{tab:planar-split-roles} records how each frozen split could influence development. Scene identifiers were mutually disjoint. Only policy training contributed supervision, gradient updates, and estimated normalization or budget scales; each later split was opened only after its preceding decision was complete.

\begin{center}
\centering
\small
\captionsetup{hypcap=false}
\captionof{table}{Prespecified roles of the planar-study data splits. ``Internal gate'' denotes a simulator-level development decision, not clinical validation.}
\label{tab:planar-split-roles}
\begin{tabularx}{\textwidth}{@{}>{\raggedright\arraybackslash}p{0.19\textwidth}>{\raggedleft\arraybackslash}p{0.07\textwidth}>{\raggedright\arraybackslash}X@{}}
\toprule
Split & Scenes & Permitted role \\
\midrule
Policy training & 448 & Teacher supervision, behaviour-cloning gradients, at most one DAgger round, and estimation of all learned normalization scales and the fixed simulated-blood margin. \\
Internal development & 64 & Teacher admission, early stopping, and development-gate rollouts; no gradient updates. Its repeated use makes it unsuitable for final performance estimation. \\
Tuning & 64 & Comparison of at most 12 prespecified behaviour-cloning configurations after the internal gate; selection of at most three configurations for validation. \\
Validation & 128 & One evaluation of at most three candidate configurations, each retrained with three frozen seeds; selection of one configuration, seed, and checkpoint by a prespecified hierarchy. \\
One-shot test & 128 & In-distribution evaluation after freezing the code, checkpoint, scales, shield, thresholds, and reporting scripts; no retraining, reselection, or retesting in response to its result. \\
Stress & 128 & Post-test out-of-distribution audit with the same frozen model under increased vascular density, cross-section, and adjacency; reported separately and never used for training or selection. \\
\bottomrule
\end{tabularx}
\end{center}

The internal development gate was a prespecified go/no-go screen applied to the teacher and then to the shielded learned controller before tuning data were opened. It required complete legal episodes, no failure or simulator-invariant violation, no per-scene excess beyond the frozen simulated-blood margin, a favourable paired time interval, and the corresponding aggregate blood criterion. For the learned controller, it additionally required retention of at least 50\% of the teacher's mean time improvement. Failure stopped the route, apart from the single DAgger round allowed after an initial learned-policy failure; passing the gate only authorized progression to tuning and did not constitute test evidence or a claim of clinical safety.

For validation selection, all three seeds of a configuration had to pass their individual gates. Eligible configurations were ordered by their three-seed mean paired time difference, mean paired blood difference, intervention rate, reported 95th-percentile wall time, and configuration identifier, in that order; the same hierarchy then selected a seed within the chosen configuration. The exact gate and selection implementation is \texttt{finalize\_validation\_v106.py}.

The v10.6 split generator \texttt{prepare\_clinical\_v106\_splits.py} uses master seed $2026081206$ and training seeds $2026081601$--$2026081603$. The separate v10.8 confirmation manifest uses master seed $2026090301$; its first 128 geometries are reused across all five sensitivity conditions. These generators and selection rules are archived at \url{https://github.com/ZJUMAI/VoxelSage/tree/e134767c/Research/planar-resection-planning}. The complete surface-scoring implementation, including scale and routing thresholds, is \url{https://github.com/ZJUMAI/VoxelSage/blob/e134767c/Port_B/skills/builtin/plan_resection/reward_function/candidate_reward.py}. The corrected sensitivity evaluator \texttt{evaluate\_v108\_sensitivity.py} freezes a separate C0 baseline for every condition; \texttt{report\_v108\_condition\_sensitivity.py} verifies the source hashes and budgets before producing the archived summary in \breakablett{artifacts/v10.8-condition-sensitivity}. Checkpoints and large generated scene files are not bundled with the source; the frozen identities and generator settings are needed in addition to the table below.

\subsection{Released checkpoint configuration}

\begin{center}
\centering
\small
\captionsetup{hypcap=false}
\captionof{table}{Behaviour-cloning training configuration for the frozen ranker. Values are those of the released checkpoint and are required for exact reproduction.}
\label{tab:bc-config}
\begin{tabularx}{\textwidth}{@{}>{\raggedright\arraybackslash}p{0.52\textwidth}>{\raggedright\arraybackslash}X@{}}
\toprule
Component & Value \\
\midrule
Hidden width / spatial channels & $96$ / $32$ \\
Encoder & $3\times$ dilated conv, dilations $1,2,4$, ReLU \\
Semantic grid channels & $10$ bit-packed + $1$ quantised transfer \\
Candidate / global context dim & $25$ / $19$ \\
Loss weights (rank, $T_{\mathrm{tot}}$, $B_{\mathrm{tail}}$, $B_{\mathrm{tot}}$, comp., safe) & $1.0, 0.2, 0.3, 0.3, 0.1, 0.2$ \\
Optimiser & AdamW, $\mathrm{lr}=3\times10^{-4}$, $\mathrm{wd}=10^{-5}$ \\
Gradient clipping & $5.0$ (global-norm) \\
Batch size / epochs & $256$ / $5$ \\
Seed & $2026081603$ \\
States / valid candidate slots / scenes & $171{,}401$ / $1{,}017{,}114$ / $448$ \\
\bottomrule
\end{tabularx}
\end{center}

\begingroup\small
\noindent\textit{Parameter meanings.}
\par\nopagebreak[4]\medskip
\noindent\begin{minipage}[t]{0.485\textwidth}
\vspace{0pt}
\begin{description}
    \item[Hidden width / spatial channels.] The hidden width of $96$ is used by the candidate embedding, global-context embedding, and shared trunk. The spatial width of $32$ is the number of feature maps produced by every convolutional layer.
    \item[Encoder.] Three $3\times3$ convolutions with dilation factors $1$, $2$, and $4$ enlarge the receptive field without reducing the $30\times40$ grid resolution; a ReLU follows each convolution.
    \item[Semantic grid channels.] Ten binary masks encode the domain, cut cells, hidden, exposed, and sealed vessel cells, frontier, large-vessel cells, current and previous positions, and start cell. They are bit-packed only for dataset storage and are unpacked before entering the network. The eleventh channel is the normalized transfer-distance map quantised to $8$ bits for storage.
    \item[Candidate / global context dimension.] Each candidate has $25$ scalar features describing its provenance, transfer and one-step action costs, vessel geometry and state, clamp-phase timing, newly exposed or sealed area, bleeding rate, and progress. The $19$ global features summarize phase and elapsed time, cutting progress, vessel counts and area, bleeding and clamping context, and the scene-specific blood-budget state.
\end{description}
\end{minipage}\hfill
\begin{minipage}[t]{0.485\textwidth}
\vspace{0pt}
\begin{description}
    \item[Loss weights.] The six coefficients multiply, in order, the teacher-choice ranking loss, full-episode time regression, post-candidate S-tail blood regression, full-episode blood regression, completion classification, and exact-safe classification. They combine offline training signals and are not an online per-action cost.
    \item[Optimiser.] AdamW updates all trainable parameters. The learning rate controls the update scale, while weight decay supplies decoupled $L_2$-style regularisation.
    \item[Gradient clipping.] Before each optimiser update, the global norm of all parameter gradients is capped at $5.0$ to limit unstable updates.
    \item[Batch size / epochs.] One update uses up to $256$ decision states, and five epochs mean five passes through the policy-training states.
    \item[Seed.] Seed $2026081603$ fixes Python, NumPy, and PyTorch random states used for parameter initialization and training-state shuffling.
    \item[States / valid candidate slots / scenes.] The dataset contains $171{,}401$ decision-state snapshots from $448$ policy-training scenes. Across those states, $1{,}017{,}114$ valid candidate entries supply ranking and auxiliary supervision; padded invalid slots are excluded from the masked losses.
\end{description}
\end{minipage}

\endgroup

\end{document}